\documentclass[sigconf]{acmart}
\AtBeginDocument{
  }

\copyrightyear{2026}
\acmYear{2026}
\setcopyright{cc}
\setcctype{by}
\acmConference[WWW '26]{Proceedings of the ACM Web Conference 2026}{April 13--17, 2026}{Dubai, United Arab Emirates}
\acmBooktitle{Proceedings of the ACM Web Conference 2026 (WWW '26), April 13--17, 2026, Dubai, United Arab Emirates}
\acmPrice{}
\acmDOI{10.1145/3774904.3792514}
\acmISBN{979-8-4007-2307-0/2026/04}

\usepackage[utf8]{inputenc} 
\usepackage[T1]{fontenc}   
\usepackage{colortbl}
\usepackage{hyperref}      
\usepackage{url}          
\usepackage{booktabs}    
\usepackage{amsfonts}      
\usepackage{nicefrac}      
\usepackage{microtype}    
\usepackage{xcolor}        
\usepackage{array} 
\usepackage{multirow}
\usepackage{tabularx}
\usepackage{graphicx}
\usepackage{enumitem}
\usepackage{amsmath}
\usepackage{subfigure}
\usepackage{wrapfig}
\usepackage{float}
\usepackage{algorithm}
\usepackage{algorithmic}
\usepackage{microtype}
\usepackage{xspace}

\newcommand{\eg}{\emph{e.g.,}\xspace}
\newcommand{\ie}{\emph{i.e.,}\xspace}

\begin{document}

\title{Adaptive Location Hierarchy Learning for Long-Tailed Mobility Prediction}

\author{Yu Wang}
\orcid{0009-0003-5193-1841}
\email{yu.wang@zju.edu.cn}
\authornote{Equal contribution}
\affiliation{
\institution{Zhejiang University}
\city{Hangzhou}
\country{China}
}

\author{Junshu Dai}
\orcid{0009-0005-0081-1799}
\authornotemark[1]
\email{djs@zju.edu.cn}
\affiliation{
  \institution{Zhejiang University}
  \city{Hangzhou}
  \country{China}
}

\author{Yuchen Ying}
\orcid{0009-0004-5683-452X}
\email{yingyc@zju.edu.cn}
\affiliation{
  \institution{Zhejiang University}
  \city{Hangzhou}
  \country{China}
}

\author{Hanyang Yuan}
\orcid{0009-0002-9570-2154}
\email{yuanhanyang@zju.edu.cn}
\affiliation{
  \institution{Zhejiang University}
  \city{Hangzhou}
  \country{China}
}

\author{Zunlei Feng}
\orcid{0000-0001-8640-8434}
\email{zunleifeng@zju.edu.cn}
\affiliation{
  \institution{Zhejiang University}
  \city{Hangzhou}
  \country{China}
}

\author{Tongya Zheng}
\orcid{0000-0003-1190-9773}
\email{doujiang_zheng@163.com}
\authornote{Corresponding author}
\authornote{Also with, High-Performance Intelligent Computing Research Center for Ultra-Large Scale Graph Data, Hangzhou City University}
\affiliation{
\institution{
Hangzhou City University, \\
Zhejiang University}
\city{Hangzhou}
\country{China}
}

\author{Mingli Song}
\orcid{0000-0003-2621-6048}
\email{brooksong@zju.edu.cn}
\affiliation{
  \institution{Zhejiang University}
  \city{Hangzhou}
  \country{China}
}

\renewcommand{\shortauthors}{Yu Wang et al.}
    
\begin{abstract}
  Human mobility prediction is crucial for applications ranging from location-based recommendations to urban planning, which aims to forecast users' next location visits based on historical trajectories.
  While existing mobility prediction models excel at capturing sequential patterns through diverse architectures for different scenarios, they are hindered by the long-tailed distribution of location visits, leading to biased predictions and limited applicability. 
  This highlights the need for a solution that enhances the long-tailed prediction capabilities of these models with broad compatibility and efficiency across diverse architectures.
  To address this need, we propose the first architecture-agnostic plugin for long-tailed human mobility prediction, named \textbf{A}daptive \textbf{LO}cation \textbf{H}ier\textbf{A}rchy learning (ALOHA).   
  Inspired by Maslow's theory of human motivation, we exploit and explore common mobility knowledge of head and tail locations derived from human mobility trajectories to effectively mitigate long-tailed bias.
  Specifically, we introduce an automatic pipeline to construct city-tailored location hierarchies based on Large Language Models (LLMs) and Chain-of-Thought (CoT) prompts, capturing high-level mobility semantics with minimal human verification. We further design an Adaptive Hierarchical Loss (AHL) that rebalances learning through Gumbel disturbance and node-wise adaptive weighting, enabling both exploitation of multi-level signals and exploration within semantically related groups.
  Extensive experiments across multiple state-of-the-art models demonstrate that ALOHA consistently improves long-tailed mobility prediction performance by up to 16.59\% while maintaining efficiency and robustness. 
  Our code is at https://github.com/Star607/ALOHA.
\end{abstract}



\maketitle

\section{Introduction}
Human mobility prediction, which forecasts an individual's next location based on historical trajectories~\cite{yang2020location}, has garnered growing attention due to the proliferation of large-scale mobility data derived from location-based social networks (LBSNs) of Web services~\cite{sanchez2022point}. 
By leveraging these data, mobility prediction advances Web applications such as location-aware recommendations, personalized content, and real-time service delivery~\cite{dai2025learning, wang2024star}, while also supporting critical urban functions including traffic management and epidemic monitoring~\cite{chen2024intensity, guo2025dual, wang2024stega}.
However, issues like privacy constraints, sparse sampling, and authorization biases often lead to a long-tailed distribution in visit frequency of locations, complicating accurate, fair and diverse human mobility prediction~\cite{xu2024taming, wang2024cola}.

Existing mobility prediction models excel at capturing sequential mobility patterns, leveraging diverse architectures for different scenarios, including Recurrent Neural Network (RNN)-based models~\cite{feng2018deepmove,sun2020go,yang2020location} for short-term and localized dependencies, Graph Neural Network (GNN)-based models~\cite{rao2022graph,yan2023spatio,yang2022getnext, wang2024star,zhang2024hyper} for spatial and semantic correlations, Transformer-based models~\cite{feng2024rotan,sun2024going,yan2023spatio,yang2022getnext} for long-range and global context, and Diffusion-based models~\cite{long2024diffusion,qin2023diffusion} for uncertain generative dynamics. However, most approaches ignore the intrinsic long-tailed distribution of location visits, causing biased predictions that degrade accuracy, fairness and recommendation diversity in applications. LoTNext~\cite{xu2024taming} is the first attempt to explicitly address this issue, but its tight coupling to graph structures limits applicability across diverse prediction architectures.

The limitations of existing methods underscore the pressing solution capable of mitigating biases from the inherently long-tailed distribution of location visits, while remaining seamlessly compatible with a wide range of state-of-the-art mobility prediction models. 
Achieving this objective entails addressing two fundamental and intertwined challenges. Firstly, how to integrate the long-tailed learning mechanism in a manner that ensures broad \textbf{\textit{compatibility}} with diverse mobility prediction architectures such as RNN-, GNN-, Transformer-, and Diffusion-based paradigms for different scenarios. 
Secondly, as broad compatibility may incur substantial computational costs, how to achieve computation \textbf{\textit{efficiency}}, enabling the solution to scale effectively and operate robustly across extensive datasets without compromising its versatility.

The challenges of long-tailed human mobility prediction in compatibility and efficiency motivate us to develop a compatible and efficient long-tailed learning plugin to complement existing sequential models, which lack explicit mechanisms for addressing long-tailed bias. Unlike tightly coupled designs that struggle to transfer across various models and datasets, this plugin integrates long-tailed learning mechanisms without modifying the core model architecture. 
Additionally, the plugin's efficiency requirement drives a lightweight design that enriches location semantics.
Drawing on Maslow's theory of human motivation~\cite{maslow1943theory}, we consider common mobility knowledge (\ie structured high-level patterns) across all locations as supplementary supervision, facilitating exploitation with high-level optimization signals for all locations and exploration through reasonable variations within higher-level groups.

In this paper, we propose \textbf{A}daptive \textbf{LO}cation \textbf{H}ier\textbf{A}rchy learning (ALOHA), a lightweight and architecture-agnostic plugin for mitigating long-tailed bias in mobility prediction. Guided by Maslow's theory of human motivation, ALOHA rebalances optimization via a hierarchical tree structure integrating coarse-grained semantics. First, we design city-tailored hierarchies with Large Language Models (LLMs) and Chain-of-Thought (CoT) prompts, capturing high-level behavioral patterns with minimal human verification. Second, given location-level logits, we directly compute hierarchical predictions via the hierarchy without additional classifiers and optimize them with Adaptive Hierarchical Loss (AHL), which incorporates (1) Gumbel disturbance to explore long-tailed locations, with its convergence property ensuring prediction accuracy, and (2) adaptive weights for node-wise adjustments across hierarchy levels. This design enables effective use of multi-level semantics and exploration of semantically related groups, ensuring compatibility and efficiency in long-tailed learning.
In conclusion, our main contributions are summarized as follows:

\begin{itemize}[itemsep=2pt,topsep=0pt,parsep=0pt,leftmargin=*]
    \item We introduce ALOHA, the first architecture-agnostic plugin for long-tailed mobility prediction, which decouples long-tailed bias mitigation from specific model architectures and ensures broad compatibility and efficiency grounded in Maslow's theory.
    \item We propose an automatic pipeline that utilizes LLMs and CoT prompts to derive city-tailored hierarchies with minimal human verification and design an adaptive hierarchical loss integrating exploitation and exploration of common mobility knowledge.
    \item ALOHA improves advanced mobility prediction methods by up to 16.59\% while demonstrating strong robustness and efficiency. In-depth analyses reveal its underlying mechanisms in detail for both head and tail location prediction.
\end{itemize}

\section{Related Work}

\noindent\textbf{Mobility Prediction.} Early work includes Markov-chain models~\cite{rendle2010factorizing, feng2015personalized} and mechanistic approaches~\cite{song2010modelling, jiang2016timegeo}, which struggle with complex spatiotemporal dependencies and rely heavily on handcrafted features. Recent studies adopt deep learning to better model sequential mobility: RNN-based methods~\cite{feng2018deepmove, sun2020go, yang2020location} capture local transitions; GNN-based models~\cite{rao2022graph, yan2023spatio, yang2022getnext, wang2024star, zhang2024hyper, xu2024taming} encode spatiotemporal relations via graphs; Transformers~\cite{feng2024rotan, sun2024going, yang2022getnext} model global dependencies with attention; and Diffusion-based approaches~\cite{long2024diffusion, qin2023diffusion} learn mobility distributions through denoising. Despite these advances, most methods remain biased toward frequently visited locations due to long-tailed visitation patterns. LoTNext~\cite{xu2024taming} explicitly addresses this but is limited by its graph-dependent design. In contrast, we introduce a lightweight, model-agnostic plugin to alleviate long-tail bias across architectures.

\noindent\textbf{Long-Tailed Learning.} 
Long-tailed learning (LTL) has been widely studied in computer vision (CV)~\cite{du2023no} and recommendation systems (RecSys)~\cite{kim2019sequential}. CV methods mainly address imbalance via data resampling~\cite{cubuk2020randaugment, shi2023re}, loss design~\cite{cui2019class, lin2017focal, park2021influence}, and logit adjustment~\cite{menonlong, tian2020posterior, wang2023balancing}. In RecSys, where sparsity exacerbates tail effects, meta-learning~\cite{wei2023meta} and transfer learning~\cite{zhang2021model} leverage shared item knowledge. However, these approaches target static class imbalance and fail to model the dynamic spatiotemporal and sequential dependencies of human mobility, limiting their applicability to mobility prediction.

\noindent\textbf{LLM for Mobility.} 
Several recent studies~\cite{he2025rhythm, du2trajagent} employ Large Language Models (LLMs)~\cite{wang2025llm4dsr,wang2025msl} to enhance mobility learning. MobilityGPT~\cite{haydari2024mobilitygpt} casts prediction as an autoregressive task with geospatial constraints and reinforcement learning. Mobility-LLM~\cite{gong2024mobility} and LLM-Mob~\cite{wang2023would} incorporate trajectories, preferences, and context to improve interpretability and accuracy. LLMob~\cite{jiawei2024large} and TrajLLM~\cite{ju2025trajllm} embed LLMs in agent-based frameworks using Chain-of-Thought reasoning~\cite{yuan2025tree} and persona generation for realistic simulation, while OpenCity~\cite{yan2024opencity} extends this paradigm to city-scale activities. Despite these advances, handling the long-tailed distribution of location visits remains largely unexplored.

\begin{figure*}[!t]
  \centering	
  \vspace{-3.5ex}
  \includegraphics[width=\textwidth]{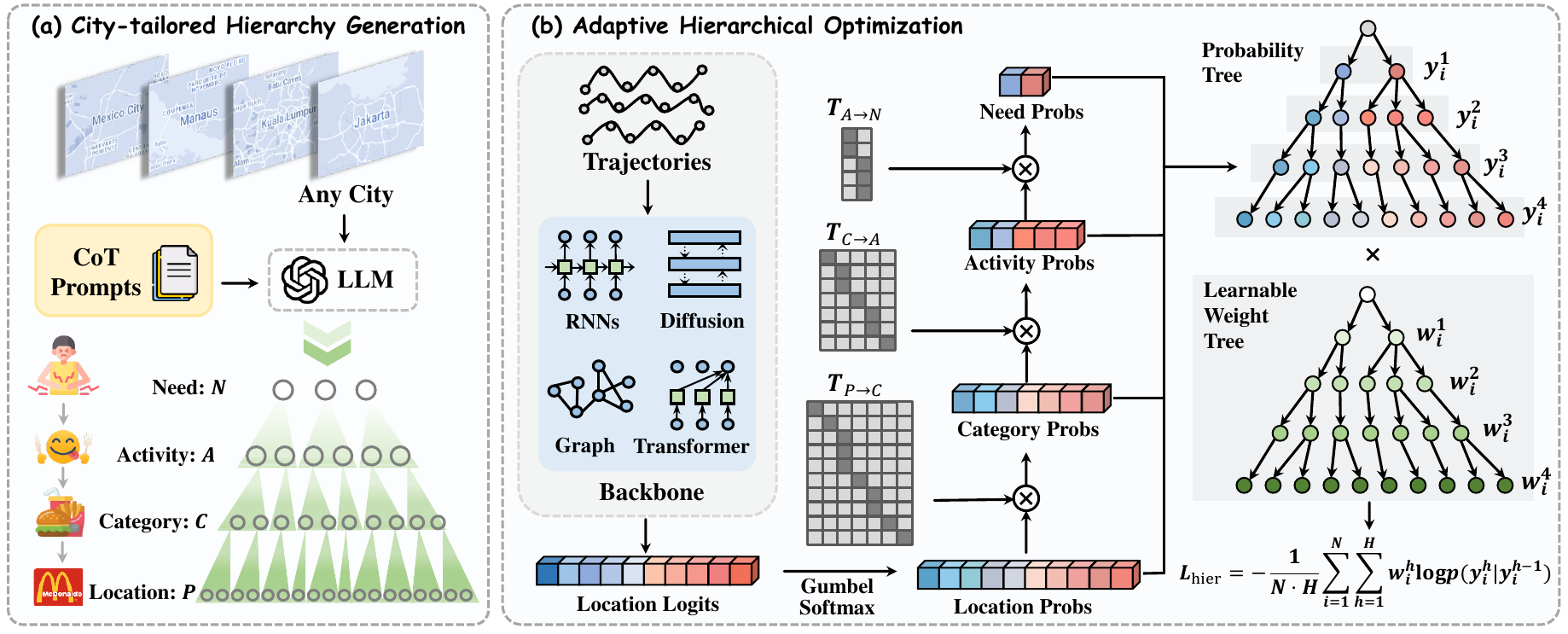}
  
  \caption{The framework of our proposed ALOHA. (a) City-tailored Hierarchy Generation: hierarchies are automatically constructed for any city using LLMs with CoT prompts, requiring minimal human verification for reliability. (b) Adaptive Hierarchical Optimization: based on constructed hierarchies, logits from arbitrary prediction architectures are transformed into probability trees via Gumbel-Softmax, with node-wise adaptive weights yielding the Adaptive Hierarchical Loss.}
  
  \label{fig:ALOHA}
  \vspace{-2.5ex}
  \end{figure*}

\section{Preliminary}

Let $ \mathcal{U}=\{u_i \mid i \in |\mathcal{U}| \}$ denote users and $\mathcal{P}=\{l_i \mid i \in |\mathcal{P}| \}$ locations, where $l_i=\langle \text{lat},\text{lon}\rangle$. $|\mathcal{U}|$ and $|\mathcal{P}|$ are sizes. A check-in $\langle u,l,cat,t\rangle$ means user $u$ visits location $l$ (category $cat$) at time $t$.

\begin{definition}[Trajectory] \label{trajectory_definition}
  A trajectory $\mathbf{x} = \{\mathbf{x}_t \mid t \in |T|\}$ is a spatiotemporal sequence where each $\mathbf{x}_t=\langle l,cat,t\rangle$ is a visit by user $u$. The set of all trajectories is denoted as $\mathcal{X} = \{\mathbf{x}^m \mid m \in M\}$
\end{definition}

\begin{definition}[Head and Tail Locations] \label{long_tail}
  Following the Pareto principle~\cite{pareto1964cours}, locations are split by visit frequency: the top 20\% form the head $\mathcal{P}_{head}$, and the rest form the tail $\mathcal{P}_{tail} = \mathcal{P} \setminus \mathcal{P}_{head}$.
\end{definition}

\begin{definition}[Mobility Prediction] \label{mobility_pred}
    Given the trajectory $\mathbf{x}_{1:t}$ of user $u$, predict a distribution over $|\mathcal{P}|$ to rank next-location candidates $\mathbf{y}_{t+1}$, focusing on improving performance on $\mathcal{P}_{tail}$.
\end{definition}

\section{Methodology}

In this section, we present the details of \textbf{A}daptive \textbf{LO}cation \textbf{H}ier\textbf{A}rchy learning (ALOHA), a lightweight, architecture-agnostic plugin for mitigating long-tailed bias in human mobility prediction. Guided by Maslow's theory of human motivation, ALOHA integrates coarse-grained semantics via a hierarchical tree. \textbf{City-tailored Hierarchy Generation} automatically constructs hierarchical labels for arbitrary cities using LLMs with designed CoT prompts, requiring minimal human effort. \textbf{Adaptive Hierarchical Optimization} enhances mobility representations through Adaptive Hierarchical Loss (AHL), combining Gumbel disturbance and node-wise adaptive weights to rebalance head and tail locations. These designs enable multi-level exploitation and exploration across semantically related locations, ensuring broad compatibility and efficiency.

\subsection{City-tailored Hierarchy Generation}

The long-tailed distribution of location visits leads models to overemphasize frequently visited locations, amplifying prediction bias. To address this, we introduce coarse-grained semantic groupings via a hierarchical tree grounded in Maslow's theory of human motivation~\cite{maslow1943theory}, rebalancing the optimization process. Behavioral psychology suggests that human mobility reflects need-driven decisions: personal needs motivate activities, which occur at specific location categories, ultimately resulting in visits to particular locations. While factors such as weather or events may also influence mobility, we focus on the need factor, as it encompasses most motivations for mobility. Accordingly, our hierarchy comprises four granularity levels: Need → Activity → Category → Location.

\subsubsection{Definition of Hierarchical Labels}
Our hierarchical structure assigns each location prediction $\mathbf{y}_{t+1}$ to multi-granularity labels $\{y^h_{t+1}\mid h \in [1, H]\}$ with $H=4$.
Each label $y^h_{t+1} \in \{1, 2, \dots, C^h \}$ denotes the class at level $h$, where $C^h$ is the number of class at this level.
Label granularity increases with $h$ (\ie $C^{h-1} < C^h$), where $C^H=\mathcal{P}$.
As outlined in Definition~\ref{trajectory_definition}, raw trajectories $\mathcal{X}$ provide location-level labels $y^{H} = \{l_i \mid i \in [1, C^{H}]\}$ and category-level labels $y^{H-1} = \{cat_i \mid i \in [1, C^{H-1}]\}$. 
Thus, our focus turns to deriving coarser-grained labels:
(1) \textbf{\textit{activity-level labels}} are derived from Foursquare\footnote[1]{\url{https://docs.foursquare.com}} with ten classes.
(2) \textbf{\textit{need-level labels}} are adapted from Maslow's theory, focusing on three fundamental human motivations: physiological needs (survival necessities such as food and shelter), safety needs (security-driven needs including education and employment), and social needs (such as recreational or social activities). Higher-order needs (esteem, self-actualization) are excluded due to their abstract, long-term nature, which lacks empirical observability in mobility data.
The proposed hierarchical structure implements Maslow's theory in a more flexible manner, departing from the original rigid order.

\subsubsection{Relations of Adjacent-level Labels}
We construct transition matrices between adjacent-level labels from bottom to top.
Each transition matrix $T_{h \rightarrow h-1}$ quantifies the likelihood of finer-grained labels at level $h$ mapping to coarser-grained labels at level $h-1$.
The location-to-category transition matrix $T_{H \rightarrow H-1}$ is directly extracted from raw trajectories $\mathcal{X}$.
For the category-to-activity transition matrix $T_{H-1 \rightarrow H-2}$, we leverage LLMs (GPT-4o mini) with designed CoT prompts (Appendix~\ref{sec:prompts}) to infer mappings from category-level labels (241\textasciitilde330 classes for different cities) to activity-level labels (10 classes applicable to all cities).
This LLM-grounded approach not only enhances mapping accuracy but also addresses the challenges posed by the variability in category definitions across different urban contexts.
Generated mappings are validated by three domain experts following established annotation protocols~\cite{stab2014annotating}, minimizing manual effort while ensuring reliability. Further details can be found in Appendix~\ref{sec:annotation}. 
The activity-to-need transition matrix $T_{H-2 \rightarrow H-3}$ follows the same derivation procedure. 
Finally, we obtain a four-tier city-tailored hierarchy, as shown in Fig.~\ref{fig:ALOHA}(a), providing structured high-level patterns to guide optimization in ALOHA.

\subsection{Adaptive Hierarchical Optimization}
Adaptive Hierarchical Optimization integrates exploitation and exploration of common mobility knowledge across head and tail locations using the city-tailored hierarchy. To exploit shared semantics, 
location-level predictions are directly extended to hierarchical predictions without extra classifiers, enabling coarse-to-fine learning.
Mobility knowledge exploration is achieved via two key components:
(1) Gumbel disturbance rebalances the learning of head and tail locations, with its convergence property ensuring prediction accuracy, and
(2) Adaptive weights adjust all nodes' importance within the hierarchy, ensuring balanced optimization. Together, these designs enable effective exploitation
of multi-level semantics and exploration within semantically related location groups.

Specifically, we extend the original mobility prediction task in Definition~\ref{mobility_pred} to learn the unknown hierarchical classification distribution based on the hierarchical labels of human motivation:
\begin{equation}
    p(\mathbf{y}_{t+1} \mid \mathbf{x}_{1:t}) = p(y^1_{t+1}, \dots, y^H_{t+1} \mid \mathbf{x}_{1:t}).
\end{equation}
Our framework first obtains the finest-grained prediction $p(y^H_{t+1} \mid \mathbf{x}_{1:t})$ using state-of-the-art mobility prediction methods with diverse backbone architectures, including RNN, GNN, Transformer and Diffusion models. 
Subsequently, the coarse-grained predictions from $y^{h-1}_{t+1}$ to $y^{h}_{t+1}$ are recursively generated via hierarchical transition matrices in a bottom-up manner, as illustrated in Fig~\ref{fig:ALOHA}(b).
This hierarchical classification pipeline offers three core advantages:
\begin{itemize}[itemsep=2pt,topsep=0pt,parsep=0pt,leftmargin=*]
    \item \textbf{\textit{Consistency}}: Directly computing coarse-grained classifications using hierarchical transition matrices eliminates noise propagation from conflicting supervisory signals.
    \item \textbf{\textit{Effectiveness}}: Our framework prioritizes fine-grained classification as the primary task, while coarser-grained predictions serve as auxiliary constraints to mitigate granularity competition\cite{chang2021your}.
    \item \textbf{\textit{Efficiency}}: Predictions at the coarse-grained level are derived directly through transition matrices, eliminating reliance on the parameters of the coarse-grained classifiers.
\end{itemize}

\subsubsection{Gumbel Disturbance.}
To mitigate the skewed optimization towards head locations in deep learning models for long-tailed mobility data, we introduce Gumbel disturbance to enhance the exploration of human mobility knowledge. By injecting randomness into the sampling process, Gumbel disturbance enables the model to consider various locations rather than relying solely on high-probability predictions. Notably, the Gumbel distribution is a family of highly stable distributions~\cite{jang2017categorical}, which preserves the original probability distribution even after noise is added.

Specifically, the fine-grained representation from any backbone architecture is mapped to logits:
\begin{equation}
    \mathbf{z} = \text{Backbone}(\mathbf{x}_{1:t}) = \{z_1, z_2, \dots, z_{|\mathcal{P}|}\},
\end{equation}
where $\text{Backbone}(\cdot)$ refers to a mobility prediction approach.
Subsequently, these logits generate probabilities of leaf nodes in the hierarchical tree using Gumbel-Softmax function:
\begin{equation}
    p_0 (y_{t+1}^H\mid \mathbf{x}_{1:t})=\operatorname{Gumbel Softmax}(z_i)=\frac{ \exp (({z_i}+g_i)/\tau)}{\sum_{j=1}^{C^H}  \exp ((z_j+g_j)/\tau)},
\end{equation}
where $i$ denotes the index of label $y_{t+1}^H$, $\tau$ is a temperature parameter and $g_i=-\log(-\log(a_i))$ is the noise sampled from $a_i \sim  \text{Uniform(0, 1)}$.
Level-wise prediction $p_0(y^{h-1}_{t+1})$ is then recursively derived in a bottom-up manner (\ie from fine to coarse granularity) using transition matrices:
\begin{equation}
    p_0(y^{h-1}_{t+1}) = T_{h \rightarrow h-1}p_0(y^h_{t+1}),
\end{equation}
where $T_{h \rightarrow h-1}$ is a transition matrix with a size of $C^h \times C^{h-1}$, the elements of $T_{h \rightarrow h-1}$ are in $\{0, 1\}$ and each row sums to one. 
In our hierarchical tree, the probability of leaf node $y^H_{t+1}$ is factorized along the path from root to leaf $y^H_{t+1}$ (Fig.~\ref{fig:ALOHA}(b)) via the chain rule~\cite{chen2022label}:
\begin{equation}
    p(y^H_{t+1}) = \prod_{h=1}^{H} p(y^h_{t+1}\mid y^{h-1}_{t+1}),
\end{equation}
where $H$ is the depth of the node and $p(y^h_{t+1}\mid y^{h-1}_{t+1})$ denotes the child-parent conditional probability, with root probability $p(y^0_{t+1})=1$. It can be expressed via leaf node probabilities as follows:
\begin{equation}
    p(y^h_{t+1}\mid y^{h-1}_{t+1})=\frac{\sum_{y^H_j\in\text{Leaves}(y_{t+1}^h)}p_0(y^H_j)}{\sum_{y^H_j\in\text{Leaves}(y^{h-1}_{t+1})}p_0(y^H_j)},
\end{equation}
where $j$ indexes $H$-level labels $C^H$ and $p(y^h_{t+1} \mid y^{h-1}_{t+1}) \in [0,1]$.

\subsubsection{Adaptive Weights.}
Due to the inherently long-tailed distribution of nodes at the location level, the distribution of high-level nodes is also skewed. To better explore dynamic mobility knowledge, we adopt node-wise adaptive weights. To prevent granularity competition~\cite{chang2021your}, we assign lower weights to the coarse-grained levels. For simplicity, the weights are initialized with \{1, 0.75, 0.5, 0.25\} from fine to coarse. We employ the Softplus activation function to ensure the positivity of the weights.
Therefore, the Adaptive Hierarchical Loss (AHL) is defined as:
\begin{equation}
\begin{aligned} L_{\text{hier}} &=-\frac{1}{ M \cdot (T-1) \cdot H} \sum_{m=1}^{M}\sum_{t=1}^{T-1}\sum_{h=1}^H {w}^h_{t+1} \log p (y_{t+1}^h\mid y_{t+1}^{h-1}) \\ &=-\frac{1}{ N  \cdot H} \sum_{i=1}^{N}\sum_{h=1}^H {w}^h_i \log p(y_i^h \mid y_i^{h-1}) \end{aligned},  
\end{equation}
where ${w}^h_i$ is a learnable weight and $N$ is the number of predictions.
Optimizing with AHL enhances per-level classification accuracy and penalizes errors in a hierarchy-aware way.

\begin{table*}[!t]
    \caption{M@$k$ and N@$k$ scores on JKT, KLP, and CA datasets. 
    GFlash denotes Graph-Flashback. 
    Best and second-best scores are marked in blue and light blue. Improv. is the relative percentage improvement.}
    \label{tab:main}
    \vspace{-1ex}
    \resizebox{0.95\textwidth}{!}{
        \begin{tabular}{l|ccccc|ccccc|ccccc}
            \toprule
            Dataset &
              \multicolumn{5}{c|}{JKT} &
              \multicolumn{5}{c|}{KLP} &
              \multicolumn{5}{c}{CA} \\ \hline
            Metric &
              M@1 &
              M@5 &
              M@10 &
              N@5 &
              N@10 &
              M@1 &
              M@5 &
              M@10 &
              N@5 &
              N@10 &
              M@1 &
              M@5 &
              M@10 &
              N@5 &
              N@10 \\ \hline
            GFlash &
              0.1916 &
              0.2791 &
              0.2909 &
              \cellcolor[HTML]{E4E8F5}0.3164 &
              \cellcolor[HTML]{E4E8F5}0.3452 &
              0.1729 &
              0.2591 &
              0.2703 &
              0.2950 &
              0.3221 &
              0.1801 &
              0.2785 &
              0.2918 &
              0.3202 &
              0.3520 \\
            +Focal &
              \cellcolor[HTML]{E4E8F5}0.1955 &
              \cellcolor[HTML]{E4E8F5}0.2800 &
              \cellcolor[HTML]{E4E8F5}0.2919 &
              0.3150 &
              0.3435 &
              0.1853 &
              0.2630 &
              0.2753 &
              0.2957 &
              0.3252 &
              0.1754 &
              0.2747 &
              0.2881 &
              0.3145 &
              0.3469 \\
            +CB &
              0.1645 &
              0.2387 &
              0.2489 &
              0.2699 &
              0.2947 &
              0.1502 &
              0.2147 &
              0.2260 &
              0.2432 &
              0.2700 &
              0.1727 &
              0.2624 &
              0.2743 &
              0.3022 &
              0.3315 \\
            +IB &
              0.1487 &
              0.2205 &
              0.2288 &
              0.2495 &
              0.2695 &
              0.1318 &
              0.1936 &
              0.2034 &
              0.2198 &
              0.2435 &
              0.1136 &
              0.1954 &
              0.2108 &
              0.2293 &
              0.2663 \\
            +LA &
              0.1721 &
              0.2474 &
              0.2580 &
              0.2783 &
              0.3042 &
              0.1522 &
              0.2219 &
              0.2321 &
              0.2516 &
              0.2764 &
              0.1551 &
              0.2506 &
              0.2632 &
              0.2929 &
              0.3232 \\
            +BLV &
              0.1874 &
              0.2781 &
              0.2897 &
              0.3151 &
              0.3437 &
              \cellcolor[HTML]{E4E8F5}0.1881 &
              \cellcolor[HTML]{E4E8F5}0.2680 &
              \cellcolor[HTML]{E4E8F5}0.2796 &
              \cellcolor[HTML]{E4E8F5}0.3027 &
              \cellcolor[HTML]{E4E8F5}0.3305 &
              \cellcolor[HTML]{E4E8F5}0.1893 &
              \cellcolor[HTML]{E4E8F5}0.2820 &
              \cellcolor[HTML]{E4E8F5}0.2962 &
              \cellcolor[HTML]{E4E8F5}0.3211 &
              \cellcolor[HTML]{E4E8F5}0.3554 \\
            +Ours &
              \cellcolor[HTML]{B4C6E7}0.2113 &
              \cellcolor[HTML]{B4C6E7}0.2946 &
              \cellcolor[HTML]{B4C6E7}0.3053 &
              \cellcolor[HTML]{B4C6E7}0.3295 &
              \cellcolor[HTML]{B4C6E7}0.3557 &
              \cellcolor[HTML]{B4C6E7}0.2045 &
              \cellcolor[HTML]{B4C6E7}0.2807 &
              \cellcolor[HTML]{B4C6E7}0.2923 &
              \cellcolor[HTML]{B4C6E7}0.3131 &
              \cellcolor[HTML]{B4C6E7}0.3414 &
              \cellcolor[HTML]{B4C6E7}0.2207 &
              \cellcolor[HTML]{B4C6E7}0.3185 &
              \cellcolor[HTML]{B4C6E7}0.3316 &
              \cellcolor[HTML]{B4C6E7}0.3588 &
              \cellcolor[HTML]{B4C6E7}0.3900 \\
            Improv. &
              8.08\% &
              5.21\% &
              4.59\% &
              4.14\% &
              3.04\% &
              8.72\% &
              4.74\% &
              4.54\% &
              3.44\% &
              3.30\% &
              16.59\% &
              12.94\% &
              11.95\% &
              11.74\% &
              9.74\% \\ \hline
            STHGCN &
              \cellcolor[HTML]{E4E8F5}0.1976 &
              0.2768 &
              0.2878 &
              0.3101 &
              0.3365 &
              \cellcolor[HTML]{E4E8F5}0.1961 &
              0.2708 &
              0.2826 &
              0.3025 &
              0.3312 &
              0.2031 &
              0.3015 &
              0.3139 &
              0.3401 &
              0.3695 \\
            +Focal &
              0.1950 &
              \cellcolor[HTML]{E4E8F5}0.2782 &
              0.2878 &
              \cellcolor[HTML]{E4E8F5}0.3124 &
              0.3358 &
              0.1829 &
              0.2602 &
              0.2719 &
              0.2927 &
              0.3207 &
              \cellcolor[HTML]{E4E8F5}0.2105 &
              \cellcolor[HTML]{E4E8F5}0.3048 &
              \cellcolor[HTML]{E4E8F5}0.3177 &
              \cellcolor[HTML]{E4E8F5}0.3433 &
              \cellcolor[HTML]{E4E8F5}0.3743 \\
            +CB &
              0.1734 &
              0.2422 &
              0.2519 &
              0.2701 &
              0.2938 &
              0.1346 &
              0.2006 &
              0.2101 &
              0.2285 &
              0.2515 &
              0.1828 &
              0.2768 &
              0.2902 &
              0.3164 &
              0.3480 \\
            +IB &
              0.1558 &
              0.2239 &
              0.2316 &
              0.2521 &
              0.2707 &
              0.0639 &
              0.1056 &
              0.1127 &
              0.1245 &
              0.1414 &
              0.1856 &
              0.2682 &
              0.2811 &
              0.3033 &
              0.3340 \\
            +LA &
              0.1708 &
              0.2424 &
              0.2518 &
              0.2724 &
              0.2953 &
              0.1198 &
              0.1942 &
              0.2067 &
              0.2264 &
              0.2563 &
              0.1708 &
              0.2719 &
              0.2855 &
              0.3141 &
              0.3467 \\
            +BLV &
              0.1966 &
              0.2773 &
              \cellcolor[HTML]{E4E8F5}0.2883 &
              0.3109 &
              \cellcolor[HTML]{E4E8F5}0.3375 &
              0.1937 &
              \cellcolor[HTML]{E4E8F5}0.2744 &
              \cellcolor[HTML]{E4E8F5}0.2858 &
              \cellcolor[HTML]{E4E8F5}0.3092 &
              \cellcolor[HTML]{E4E8F5}0.3367 &
              0.2059 &
              0.3001 &
              0.3127 &
              0.3395 &
              0.3698 \\
            +Ours &
              \cellcolor[HTML]{B4C6E7}0.2097 &
              \cellcolor[HTML]{B4C6E7}0.2899 &
              \cellcolor[HTML]{B4C6E7}0.3002 &
              \cellcolor[HTML]{B4C6E7}0.3239 &
              \cellcolor[HTML]{B4C6E7}0.3487 &
              \cellcolor[HTML]{B4C6E7}0.2065 &
              \cellcolor[HTML]{B4C6E7}0.2836 &
              \cellcolor[HTML]{B4C6E7}0.2946 &
              \cellcolor[HTML]{B4C6E7}0.3163 &
              \cellcolor[HTML]{B4C6E7}0.3431 &
              \cellcolor[HTML]{B4C6E7}0.2207 &
              \cellcolor[HTML]{B4C6E7}0.3145 &
              \cellcolor[HTML]{B4C6E7}0.3250 &
              \cellcolor[HTML]{B4C6E7}0.3536 &
              \cellcolor[HTML]{B4C6E7}0.3792 \\
            Improv. &
              6.12\% &
              4.21\% &
              4.13\% &
              3.68\% &
              3.32\% &
              5.30\% &
              3.35\% &
              3.08\% &
              2.30\% &
              1.90\% &
              4.85\% &
              3.18\% &
              2.30\% &
              3.00\% &
              1.31\% \\ \hline
            MCLP &
              \cellcolor[HTML]{E4E8F5}0.1968 &
              \cellcolor[HTML]{E4E8F5}0.2796 &
              \cellcolor[HTML]{E4E8F5}0.2908 &
              \cellcolor[HTML]{E4E8F5}0.3131 &
              \cellcolor[HTML]{E4E8F5}0.3400 &
              0.1833 &
              0.2594 &
              0.2703 &
              0.2909 &
              0.3174 &
              0.1939 &
              0.2905 &
              0.3016 &
              0.3285 &
              0.3551 \\
            +Focal &
              0.1961 &
              0.2771 &
              0.2881 &
              0.3102 &
              0.3364 &
              0.1713 &
              0.2508 &
              0.2616 &
              0.2830 &
              0.3094 &
              0.1911 &
              0.2871 &
              0.2985 &
              0.3259 &
              0.3530 \\
            +CB &
              0.1792 &
              0.2517 &
              0.2621 &
              0.2822 &
              0.3073 &
              0.1554 &
              0.2237 &
              0.2336 &
              0.2523 &
              0.2763 &
              0.1745 &
              0.2666 &
              0.2771 &
              0.3057 &
              0.3308 \\
            +IB &
              0.1839 &
              0.2559 &
              0.2649 &
              0.2859 &
              0.3076 &
              0.1534 &
              0.2171 &
              0.2247 &
              0.2439 &
              0.2626 &
              0.0766 &
              0.1005 &
              0.1044 &
              0.1106 &
              0.1197 \\
            +LA &
              0.1834 &
              0.2577 &
              0.2679 &
              0.2892 &
              0.3140 &
              0.1554 &
              0.2258 &
              0.2362 &
              0.2547 &
              0.2800 &
              0.1930 &
              0.2801 &
              0.2914 &
              0.3169 &
              0.3441 \\
            +BLV &
              0.1918 &
              0.2763 &
              0.2868 &
              0.3101 &
              0.3357 &
              \cellcolor[HTML]{E4E8F5}0.1885 &
              \cellcolor[HTML]{E4E8F5}0.2627 &
              \cellcolor[HTML]{E4E8F5}0.2725 &
              \cellcolor[HTML]{E4E8F5}0.2946 &
              \cellcolor[HTML]{E4E8F5}0.3186 &
              \cellcolor[HTML]{E4E8F5}0.2105 &
              \cellcolor[HTML]{E4E8F5}0.2969 &
              \cellcolor[HTML]{E4E8F5}0.3078 &
              \cellcolor[HTML]{E4E8F5}0.3329 &
              \cellcolor[HTML]{E4E8F5}0.3588 \\
            +Ours &
              \cellcolor[HTML]{B4C6E7}0.2079 &
              \cellcolor[HTML]{B4C6E7}0.2903 &
              \cellcolor[HTML]{B4C6E7}0.3015 &
              \cellcolor[HTML]{B4C6E7}0.3235 &
              \cellcolor[HTML]{B4C6E7}0.3507 &
              \cellcolor[HTML]{B4C6E7}0.1993 &
              \cellcolor[HTML]{B4C6E7}0.2715 &
              \cellcolor[HTML]{B4C6E7}0.2829 &
              \cellcolor[HTML]{B4C6E7}0.3016 &
              \cellcolor[HTML]{B4C6E7}0.3295 &
              \cellcolor[HTML]{B4C6E7}0.2225 &
              \cellcolor[HTML]{B4C6E7}0.3215 &
              \cellcolor[HTML]{B4C6E7}0.3308 &
              \cellcolor[HTML]{B4C6E7}0.3608 &
              \cellcolor[HTML]{B4C6E7}0.3830 \\
            Improv. &
              6.02\% &
              4.76\% &
              4.65\% &
              4.29\% &
              4.25\% &
              5.73\% &
              3.35\% &
              3.82\% &
              2.38\% &
              3.42\% &
              5.70\% &
              8.29\% &
              7.47\% &
              8.38\% &
              6.74\% \\ \hline
            Diff-POI &
              0.1234 &
              0.1826 &
              0.1912 &
              0.2070 &
              0.2273 &
              \cellcolor[HTML]{E4E8F5}0.1250 &
              0.1762 &
              0.1843 &
              0.1973 &
              0.2170 &
              0.1339 &
              \cellcolor[HTML]{E4E8F5}0.1932 &
              \cellcolor[HTML]{E4E8F5}0.2004 &
              \cellcolor[HTML]{E4E8F5}0.2180 &
              \cellcolor[HTML]{E4E8F5}0.2354 \\
            +Focal &
              0.1287 &
              \cellcolor[HTML]{E4E8F5}0.1865 &
              \cellcolor[HTML]{E4E8F5}0.1955 &
              \cellcolor[HTML]{E4E8F5}0.2105 &
              \cellcolor[HTML]{E4E8F5}0.2321 &
              0.1242 &
              \cellcolor[HTML]{E4E8F5}0.1794 &
              \cellcolor[HTML]{E4E8F5}0.1875 &
              \cellcolor[HTML]{E4E8F5}0.2021 &
              \cellcolor[HTML]{E4E8F5}0.2214 &
              0.1274 &
              0.1785 &
              0.1861 &
              0.2010 &
              0.2198 \\
            +CB &
              0.1282 &
              0.1826 &
              0.1876 &
              0.2046 &
              0.2167 &
              0.1078 &
              0.1605 &
              0.1650 &
              0.1814 &
              0.1922 &
              0.0960 &
              0.1487 &
              0.1557 &
              0.1723 &
              0.1893 \\
            +IB &
              0.0824 &
              0.0981 &
              0.1007 &
              0.1049 &
              0.1112 &
              0.1166 &
              0.1378 &
              0.1401 &
              0.1466 &
              0.1522 &
              0.0914 &
              0.1082 &
              0.1106 &
              0.1151 &
              0.1210 \\
            +LA &
              0.1147 &
              0.1677 &
              0.1748 &
              0.1903 &
              0.2072 &
              0.0998 &
              0.1511 &
              0.1559 &
              0.1723 &
              0.1841 &
              0.0868 &
              0.1453 &
              0.1547 &
              0.1711 &
              0.1932 \\
            +BLV &
              \cellcolor[HTML]{E4E8F5}0.1289 &
              0.1843 &
              0.1928 &
              0.2087 &
              0.2289 &
              0.1194 &
              0.1707 &
              0.1785 &
              0.1920 &
              0.2109 &
              \cellcolor[HTML]{E4E8F5}0.1431 &
              0.1912 &
              0.1996 &
              0.2127 &
              0.2330 \\
            +Ours &
              \cellcolor[HTML]{B4C6E7}0.1429 &
              \cellcolor[HTML]{B4C6E7}0.1948 &
              \cellcolor[HTML]{B4C6E7}0.2025 &
              \cellcolor[HTML]{B4C6E7}0.2175 &
              \cellcolor[HTML]{B4C6E7}0.2364 &
              \cellcolor[HTML]{B4C6E7}0.1410 &
              \cellcolor[HTML]{B4C6E7}0.1953 &
              \cellcolor[HTML]{B4C6E7}0.2020 &
              \cellcolor[HTML]{B4C6E7}0.2176 &
              \cellcolor[HTML]{B4C6E7}0.2341 &
              \cellcolor[HTML]{B4C6E7}0.1496 &
              \cellcolor[HTML]{B4C6E7}0.2022 &
              \cellcolor[HTML]{B4C6E7}0.2107 &
              \cellcolor[HTML]{B4C6E7}0.2254 &
              \cellcolor[HTML]{B4C6E7}0.2458 \\
            Improv. &
              10.86\% &
              5.70\% &
              5.03\% &
              3.33\% &
              1.85\% &
              12.80\% &
              8.86\% &
              7.73\% &
              7.67\% &
              5.74\% &
              4.54\% &
              4.66\% &
              5.14\% &
              3.39\% &
              4.42\% \\ \hline
            MELT &
              0.1388 &
              0.2276 &
              0.2399 &
              0.2632 &
              0.2936 &
              0.1139 &
              0.1962 &
              0.2072 &
              0.2314 &
              0.2573 &
              0.1241 &
              0.2016 &
              0.2128 &
              0.2353 &
              0.2631 \\
            LoTNext &
              0.1834 &
              0.2577 &
              0.2686 &
              0.2895 &
              0.3157 &
              0.1861 &
              0.2514 &
              0.2599 &
              0.2795 &
              0.3003 &
              0.2041 &
              0.2896 &
              0.3034 &
              0.3254 &
              0.3583 \\ \bottomrule
            \end{tabular}
    }
\end{table*}

\subsection{Gradient Comparison}
We compare the gradient of AHL and standard cross-entropy loss to analyse their optimization dynamics. 
For simplicity, we consider optimizing a single label $y_{t+1}^h$ rather than all samples from trajectory sequences $\mathcal{X}$. 
The optimization objectives for both losses are as:
\begin{equation}
    \ell_{\text{ce}}=-\log p(y^H_i).
\end{equation}
\begin{equation}
    \ell_{\text{hier}} = -\sum_{h=1}^H {w}^h_i \log p(y_i^h\mid y_i^{h-1}).
\end{equation}
Let $S_i^h$ be the aggregated probability of leaf nodes under $y_i^h$:
\begin{equation}
    p(y^h_i \mid y^{h-1}_i)=\frac{S_i^h}{S_i^{h-1}}, \quad S_i^h=\sum_{y^H_k \in \text { Leaves }(y^h_i)} p_0(y^H_k),
\end{equation}
The gradients of these two losses with respect to logits $z_j$ are:
\begin{equation}
    \frac{\partial \ell_\text{ce}}{\partial z_j} = p_0(y_i^H)-1.
\end{equation}
\begin{equation}
    \frac{\partial \ell_{\text{hier}}}{\partial z_j} = p_0(y^H_i) \left[ \sum_{h=1}^{H-1} \frac{{w}^{h+1}_i - {w}^h_i}{S_i^h} + {w}^1_i \right] - {w}^H_i = p_0(y^H_i) \cdot A - {w}^H_i,
\end{equation}
where the superscript of $H$ refers to the location level and $1$ represents the need level. 
It can be observed that AHL encapsulates intra-hierarchy learning of locations $({w}^{h+1}_i-{w}^h_i)/{S_i^h}$ in a competitive manner, 
implying that uneven adjustments across levels are essential to maintain representational diversity; otherwise, equal weights lead to a degraded optimization of $p_0(y^H_i){w}^1_i - {w}^H_i$.
Besides, ${w}^1_i$ and ${w}^H_i$ serve to stabilize coefficients, preventing gradient vanishing during optimization.
The progressive decrease of weight initialization across levels corresponds to a tendency towards shared sematic learning among long-tailed locations.

\vspace{-1.8ex} 
\section{Experiments}
\label{sec:experiment} 
In this section, we evaluate ALOHA's performance validate extensive experiments. More details are provided in Appendix~\ref{sec:exp_appendix}.

\subsection{Experimental Setup}
\subsubsection{Datasets.}
\label{sec:datasets}
We conduct extensive experiments on six real-world mobility datasets, sourced from Foursquare\footnote{\url{https://sites.google.com/site/yangdingqi/home/foursquare-dataset}} and Gowalla\footnote{\url{https://snap.stanford.edu/data/loc-gowalla.html}}, where locations with fewer than 15 visits and users with fewer than 100 check-ins are excluded to and ensure dataset quality~\cite{xu2024taming,yan2023spatio,qin2023diffusion}. 
Each user's check-in sequence is segmented into trajectories using 24-hour intervals. 
We sort check-ins chronologically for each user and partition the dataset into training, validation, and test sets in an 8:1:1 ratio~\cite{yang2022getnext}. 
Users or locations absent from the training set but present in validation/test sets are excluded during evaluation. 
The statistics of processed datasets are summarized in Table~\ref{tab:statistic-data}.
Further details are available in Appendix~\ref{sec:statistics}.

\begin{table}[!h]
    \centering
    \small
    \vspace{-2ex}
    \caption{Basic statistics of processed datasets.}
    \vspace{-2.3ex}
    \resizebox{0.35\textwidth}{!}{ 
        \begin{tabular}{@{}cccccc@{}} 
            \toprule
            \textbf{City} & \textbf{\#User} & \textbf{\#Loc.} & \textbf{\#Record} & \textbf{\#Traj.} & \textbf{\#Cat.} \\
            \midrule
            JKT    & 2,827  & 7,240  & 240,914 & 40,733  & 308   \\
            KLP    & 1,528  & 3,836  & 139,391 & 24,015  & 259   \\
            CA     & 1,267  & 4,171  & 94,856  & 13,217  & 260   \\
            MAO    & 1,036  & 3,180  & 123,861 & 18,735  & 241   \\
            MOW    & 3,234  & 8,963  & 316,576 & 50,702  & 330   \\
            SPB    & 992    & 3,077  & 99,832  & 15,263  & 253   \\
            \bottomrule
        \end{tabular}
    }
    \vspace{-3ex}
    \label{tab:statistic-data}
\end{table}

\subsubsection{Baselines.}
We implement four types of backbone architectures for human mobility prediction to validate our framework's generalizability: RNN-based model (Graph-Flashback~\cite{rao2022graph}), GNN-based model (STHGCN~\cite{yan2023spatio}), Transfomer-based model (MCLP~\cite{sun2024going}) and Diffusion-based model (Diff-POI~\cite{qin2023diffusion}).
We compare ALOHA's performance against two categories of state-of-the-art baselines: (1) widely adopted Long-tailed Learning (LTL) \textbf{plugins} in computer vision, including loss function (Focal~\cite{lin2017focal}, CB~\cite{cui2019class} and IB~\cite{park2021influence}) and logit adjustment (LA~\cite{menonlong} and BLV~\cite{wang2023balancing}).
(2) significant \textbf{frameworks} for long-tailed recommendation systems (MELT~\cite{kim2023melt}) and human mobility (LoTNext~\cite{xu2024taming}). 
More details see Appendix~\ref{sec:baseline}.

\subsubsection{Evaluation Metrics.} 
We evaluate model performance using MRR@$k$ and NDCG@$k$ (M@$k$ and N@$k$ for brevity) with $k \in \{1,5,10\}$. M@$k$ measures the reciprocal rank of the ground-truth label within the top-$k$ predictions, while N@$k$ discounts lower positions to emphasize accurate placement of relevant labels at higher ranks. Note that M@1 and N@1 are equivalent. 

\vspace{-1.3ex}
\subsection{Prediction Performance}
\subsubsection{Overall Comparison}
We compare the architecture-agnostic ALOHA against LTL plugin and framework baselines using M@$k$ and N@$k$ metrics across six datasets for mobility prediction. 
Partial results are presented in Table~\ref{tab:main}, and the complete results are provided in Appendix~\ref{sec:main_table_appendix}.
We can yield three key observations:
\begin{itemize}[itemsep=2pt,topsep=0pt,parsep=0pt,leftmargin=*]
    \item \textbf{Consistent Effectiveness of ALOHA.}
    ALOHA consistently achieves the highest M@$k$ and N@$k$ scores on all backbone architectures. 
    Specifically, it achieves 16.59\% improvement on RNN-based backbone (GFlash) for CA dataset and 12.80\% on Diffusion-based backbone (Diff-POI) for KLP dataset on M@1 metric. 
    This superiority arises from ALOHA's adaptive hierarchical optimization, which mitigates skewed location distributions through shared semantic among locations.
    Notably, STHGCN is the strongest backbone without plugins on CA dataset, but adding ALOHA enables the lighter GFlash to surpass ``STHGCN + Ours'', demonstrating ALOHA's effectiveness and potential.

    \item \textbf{Performance of Plugin Baselines.}
    Although some LTL plugin baselines achieve suboptimal performance on partial metrics (\eg MCLP + BLV on KLP dataset), their results are generally unstable and may even harm backbone models (\eg a 33.23\% drop on M@1 for JKT dataset with Diff-POI + IB). This is because LTL plugins, designed for computer vision (CV) data with fewer classes and extreme imbalance, tend to over-correct on locations. Moreover, independently and identically distributed CV data fail to capture the complex spatiotemporal dependencies of locations in mobility sequences, degrading overall performance.

    \item \textbf{Performance of Framework Baselines.}
    MELT underperforms LoTNext by up to 24.32\% on M@1 of KLP dataset. This discrepancy arises from the direct transplantation of LTL strategies for RecSys, which fails to address the unique challenges of mobility prediction such as varying scales of distributions, distinct prediction strategies, and the absence of spatial considerations.  
    Although LoTNext outperforms methods based on Diff-POI, it underperforms compared to those utilizing other backbones, highlighting a need for improved generalizability and flexibility.
\end{itemize}

\subsubsection{Long-Tail Location Comparison} 
\begin{figure}[htbp]
    \centering	
    \vspace{-1.5ex}
    \includegraphics[width=0.47\textwidth]{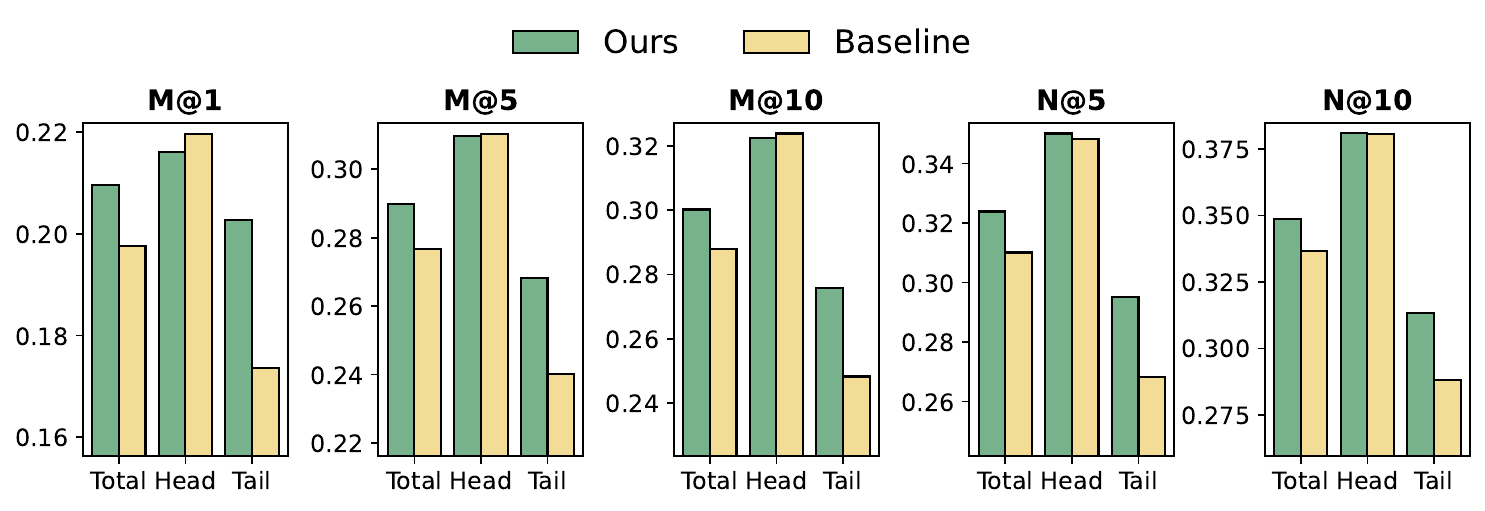}
    \caption{Prediction comparison of total/head/tail groups between ALOHA and SOTA baseline on JKT dataset.}
    \vspace{-1.5ex}
    \label{fig:head_tail}
    \vspace{-2ex}                  
\end{figure}

According to the partitioning process described in Definition~\ref{long_tail}, locations are divided into tail and head groups. 
To assess whether our model enhances performance on long-tailed locations, we compare ALOHA with the optimal baseline method (STHGCN) using the GNN-based backbone architecture on JKT dataset. Figure~\ref{fig:head_tail} demonstrates that ALOHA exhibits significant improvements in total and tail groups. 
In the head group, performance remains comparable to the baseline at M@$k$ while achieving marginally superior results at N@$k$. 
This outcome indicates that our adaptive hierarchical learning framework effectively enhances the learning process for long-tailed locations without compromising performance on head locations.

\subsection{Robustness}

\subsubsection{Robustness in Hierarchy Construction}

Based on hierarchical labels guided by Maslow's theory of human motivation, ALOHA leverages LLMs and designed CoT prompts to generate inter-level mappings, get city-tailored hierarchies. To evaluate the stability of these hierarchies, we compare results from two types of LLMs: GPT-4o-mini~\cite{hurst2024gpt} and GLM-4.5~\cite{zeng2025glm}. As shown in Table~\ref{tab:llm_comparison}, the manual correction rate for category $\rightarrow$ activity mappings is below 2.60\% across both models, indicating high consistency. This confirms the robustness of the hierarchy construction process, with minimal manual refinement required for reliability.

\begin{table}[h]
    \centering
    \vspace{-1ex} 
    \caption{Manual correction rates of hierarchical mappings generated by different LLMs on JKT and KLP datasets.}
    \vspace{-1ex} 
    \resizebox{0.38\textwidth}{!}{
        \begin{tabular}{c|c|c|c|c}
            \toprule
            \textbf{Mapping}  & \textbf{LLM}      & \textbf{JKT}         & \textbf{KLP}      & \textbf{CA}   \\ 
            \midrule
            Activity → Need     & GPT-4o-mini    & \multicolumn{3}{c}{20\% (2/10)}       \\ 
                                         & GLM-4.5        & \multicolumn{3}{c}{10\% (1/10)}       \\ 
            \midrule
            Category → Activity & GPT-4o-mini    & 0.97\%            & 1.93\%         &  0.77\% \\ 
                                         & GLM-4.5        & 2.60\%            & 2.32\%         & 1.92\%   \\ 
            \bottomrule
        \end{tabular}
    }
    \vspace{-2ex} 
    \label{tab:llm_comparison}
\end{table}

\subsubsection{Robustness to Noisy Hierarchy}
We further evaluate the impact of noisy hierarchical mappings on ALOHA's performance by randomly corrupting 10\% of the mappings for activity $\rightarrow$ need (N), category $\rightarrow$ activity (A), and both (NA). Table~\ref{tab:noisy_hierarchy} presents the results on JKT dataset, where performance remains nearly identical to the clean setting, with only minor metric variations. This demonstrates that ALOHA is robust to moderate noise in the hierarchical mappings, with its adaptive mechanisms effectively mitigating the noise's impact. 
See Appendix~\ref{sec:noisy_appendix} for results on other datasets.

\begin{table}[!h]
  \vspace{-1.8ex}
  \caption{Prediction performance under noisy hierarchical mappings on JKT dataset. ``N'', ``A'', and ``NA'' refer to noise in activity $\rightarrow$ need, category $\rightarrow$ activity, and both.}
  \vspace{-1.8ex}
  \resizebox{0.36\textwidth}{!}{
  \begin{tabular}{l|ccccc}
      \toprule
      \textbf{Metric} & \textbf{M@1} & \textbf{M@5} & \textbf{M@10} & \textbf{N@5} & \textbf{N@10} \\
      \midrule
      N               & 0.2118       & 0.2903       & 0.3014        & 0.3229       & 0.3499        \\
      A               & 0.2003       & 0.2832       & 0.2933        & 0.3172       & 0.3417        \\
      NA              & 0.2039       & 0.2870       & 0.2977        & 0.3215       & 0.3477        \\
      Ours            & 0.2097       & 0.2899       & 0.3002        & 0.3239       & 0.3487        \\ 
      \bottomrule
  \end{tabular}
  }
  \vspace{-3ex}
  \label{tab:noisy_hierarchy}
\end{table}

\subsection{Efficiency}

\subsubsection{Cost for hierarchy construction.}
We evaluate the efficiency of LLM-assisted hierarchy construction by reporting the time and cost for generating city-tailored hierarchies. As illustrated in Table~\ref{tab:cost_llm}, activity $\rightarrow$ need mappings remain consistent across cities, facilitating the reuse of JKT results. Category $\rightarrow$ activity mappings requires 3~6 minutes per city and incurs a cost of less than \$1, demonstrating that LLM-assisted mapping is both cost-effective and time-efficient.

\begin{table}[h]
    \centering
    \caption{Cost and time for hierarchy construction using LLMs (GPT-4o-mini) across six datasets.}
    \vspace{-1.5ex}
    \resizebox{0.38\textwidth}{!}{
    \begin{tabular}{lcccccc}
        \toprule
        \textbf{Dataset} & \textbf{JKT} & \textbf{KLP} & \textbf{CA} & \textbf{MAO} & \textbf{MOW} & \textbf{SPB} \\
        \hline
        \multicolumn{1}{l}{} & \multicolumn{6}{c}{Activity $\rightarrow$ Need} \\
        \hline
        Cost (\$) & \multicolumn{6}{c}{0.03} \\
        Time (min) & \multicolumn{6}{c}{0.33} \\
        \hline
        \multicolumn{1}{l}{} & \multicolumn{6}{c}{Category $\rightarrow$ Activity} \\
        \hline
        Cost (\$) & 0.86 & 0.70 & 0.74 & 0.67 & 0.94 & 0.69 \\
        Time (min) & 5.28 & 4.25 & 3.13 & 3.98 & 6.05 & 4.03 \\
        \bottomrule
    \end{tabular}
    }
    \vspace{-2ex}
    \label{tab:cost_llm}
\end{table}

\subsubsection{Computational Cost}
Table~\ref{tab:cost} presents the computational efficiency of ALOHA and plugin baselines across four backbone architectures. ALOHA introduces minimal additional parameters, with a complexity of $O(N)$, resulting in negligible overhead. Among the plugins, BLV is the slowest due to normal distribution generation, while Base, Focal, and CB exhibit similar speeds. MCLP is the fastest due to its simplicity, while Graph-Flashback's $O(N^2)$ mechanism and STHGCN's multi-hop sampling add extra overhead.

\begin{table}[hb] 
    \vspace{-1ex}
    \centering
    \caption{The average computation time of each trajectory of all plugin baselines across four backbone architectures on JKT dataset, measured in units of $10^{-3}$ second.}     
    \vspace{-1.5ex}
    \resizebox{0.39\textwidth}{!}{
  \begin{tabular}{l|cccc}
    \toprule
    Model   & Graph-Flashback & STHGCN & MCLP & Diff-POI \\
    \midrule
    Base  & 0.628           & 8.492  & 0.437 & 12.753   \\
    + Focal & 0.709           & 8.782  & 0.447 & 12.950    \\
    + CB    & 0.644           & 8.613  & 0.466 & 13.267   \\
    + IB    & 0.709           & 8.534  & 0.521 & 12.662   \\
    + LA    & 0.727           & 8.480  & 0.544 & 12.708   \\
    + BLV   & 4.978           & 8.969  & 4.918 & 13.879   \\
    \midrule
    + Ours  & 0.735           & 8.824  & 0.514 & 12.889  \\
    \bottomrule
  \end{tabular}
  }
    \label{tab:cost}
    \vspace{-2ex}
\end{table}

\subsection{Hierarchical Analysis of ALOHA}

\subsubsection{Hierarchical Distance Analysis.} 
\begin{figure}[!ht]
    \centering	
    \vspace{-1.5ex}
    \resizebox{0.45\textwidth}{!}{ 
        \includegraphics{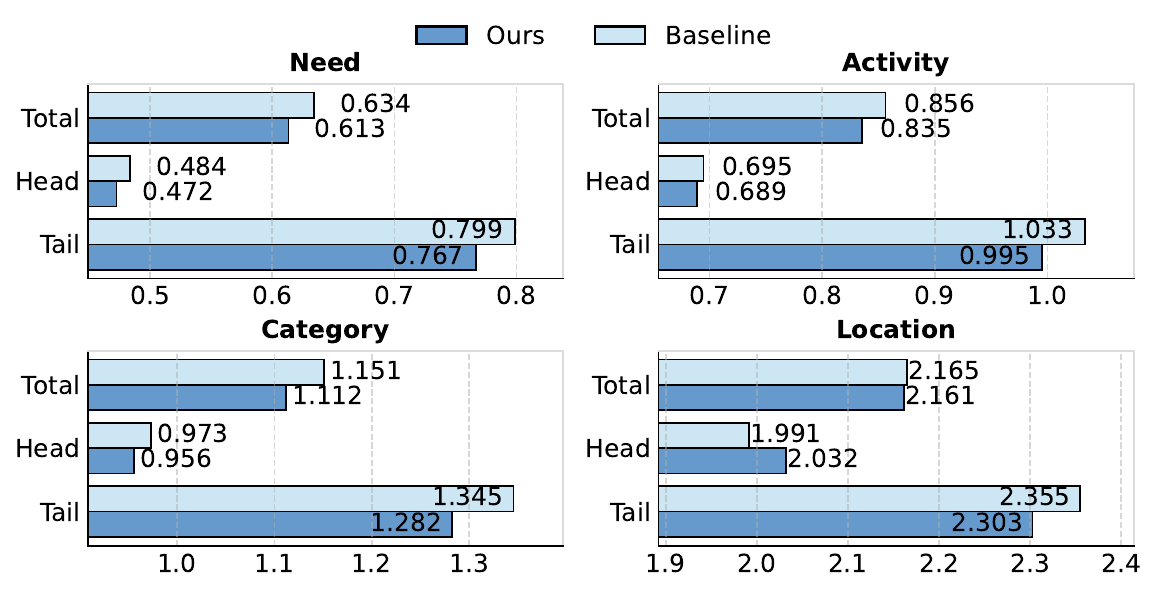}
    }
    \caption{Overall comparison of hierarchical distance between ALOHA and the SOTA baseline on the JKT dataset.}
    \label{fig:hier_dist}
    \vspace{-1.5ex}
\end{figure}

We evaluate prediction similarity based on the label hierarchy tree, computed using the lowest common ancestor between the ground truth and the predicted location. A smaller hierarchical distance indicates fewer prediction errors. Comparing ALOHA with the best baseline (STHGCN) on the JKT dataset, Figure~\ref{fig:hier_dist} shows that ALOHA achieves smaller hierarchical distances at the coarse-grained levels of need, activity, and category, with improvements of up to 3.3\%, 2.5\%, and 3.4\%, respectively. This demonstrates ALOHA's better understanding of common mobility semantics, enhancing its ability to predict long-tailed locations and improving overall performance.

\subsubsection{Hierarchical Weights Visualization.} 

\begin{figure}[!hb]
    \centering
    \includegraphics[width=0.35\textwidth]{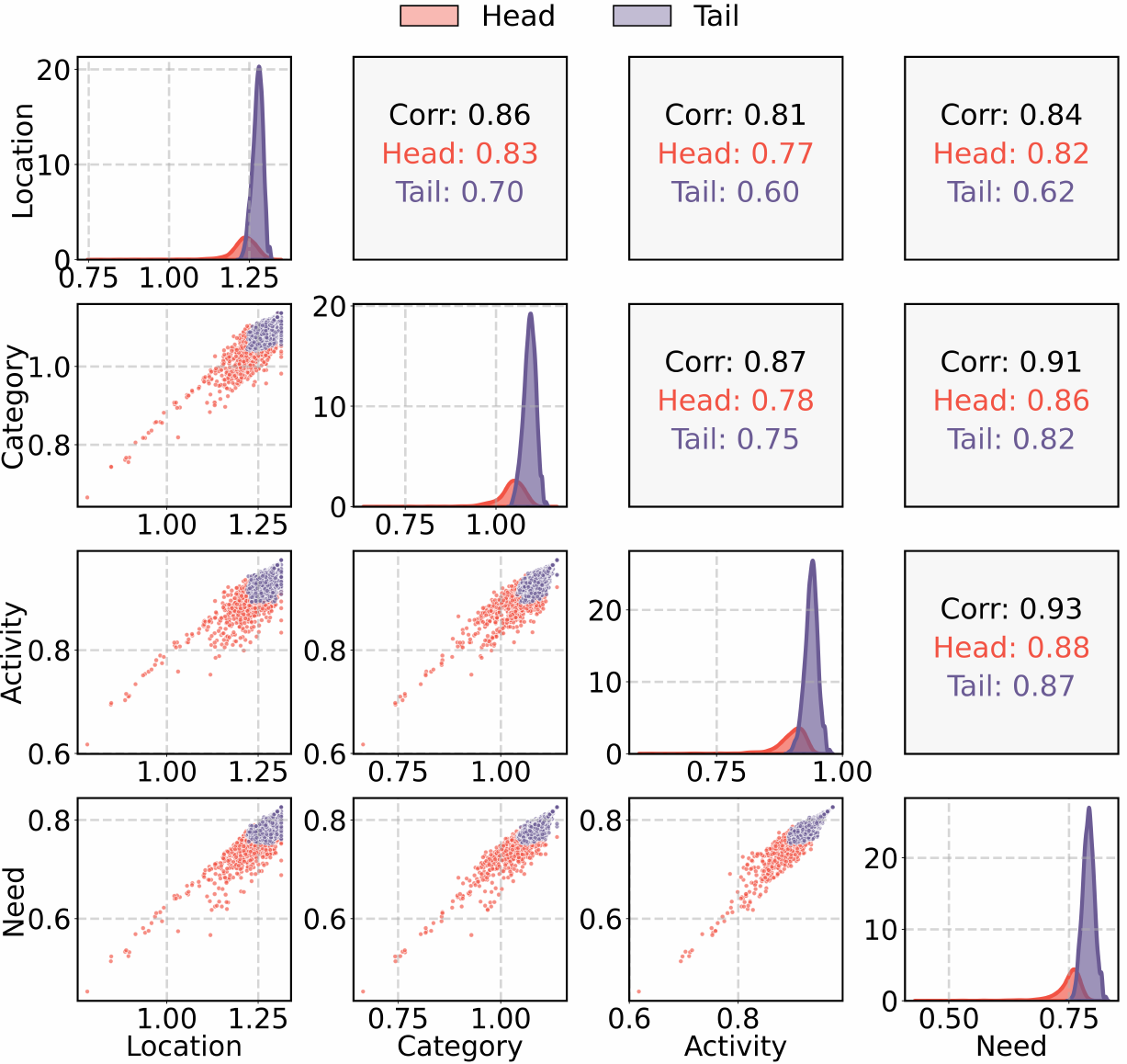}
    \caption{Optimized hierarchical weights of ALOHA on STHGCN over JKT dataset. The diagonal plots show kernel density estimates of weight distributions for head and tail groups. The lower triangular matrix displays scatterplots of hierarchical weights, while the upper triangular matrix shows correlation coefficients for all/head/tail locations.}
    \label{fig:weight}
    \vspace{-2.8ex}  
\end{figure}

We visualize the optimized weights of ALOHA on STHGCN over the JKT dataset to gain deeper insights into its adaptive mechanisms. Figure~\ref{fig:weight} presents the scatterplot matrix of the weights across four hierarchical levels. The results show that long-tailed classes exhibit a more concentrated weight distribution, with significantly larger values compared to head classes, confirming the model's adjustments for both groups. Additionally, the calculated correlation coefficients reveal stronger correlations between adjacent hierarchical levels, further demonstrating the hierarchical structure's effectiveness. In contrast, long-tailed classes display lower correlation coefficients, reflecting their more diverse characteristics. These findings underscore the model's ability to learn effectively from long-tailed locations while leveraging the interrelatedness of head locations, ultimately leading to more balanced performance in mobility prediction.

\begin{figure*}[!t]
  \centering	
  \resizebox{0.99\textwidth}{!}{ 
      \includegraphics{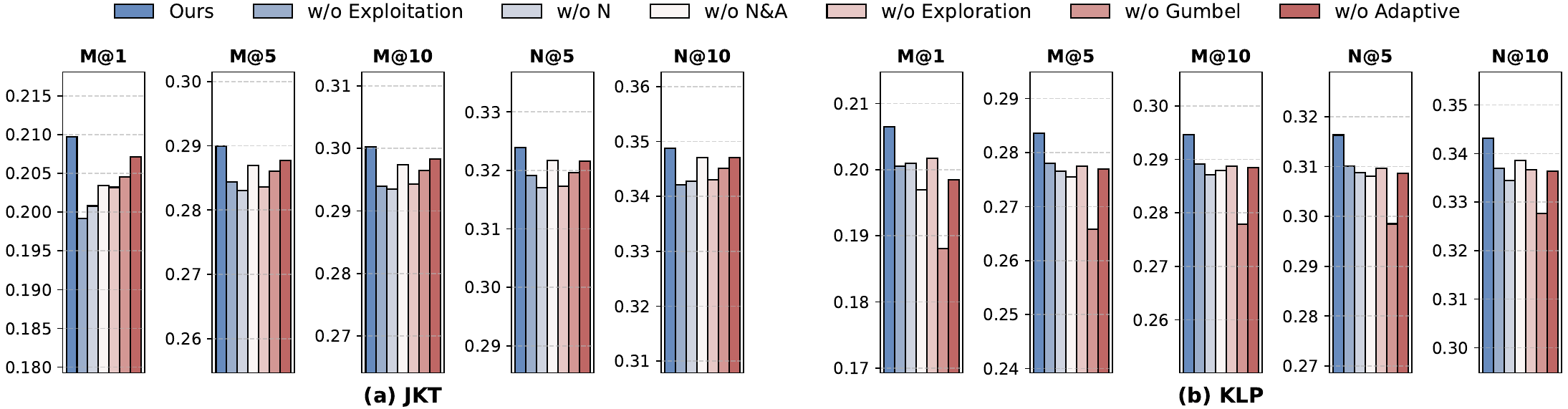}
  }
  \vspace{-1.5ex} 
  \caption{Ablation study of our proposed ALOHA using STHGCN backbone on JKT and KLP dataset.}
  \label{fig:ablation}
\end{figure*}

\begin{table*}[htbp]
  \renewcommand{\arraystretch}{1.1}
  \centering
  \caption{Case study on long-tailed location prediction. For each hierarchical level, we show the ground truth label and the top-5 predictions from both ALOHA (Ours) and the SOTA baseline (STHGCN). The columns "No.1" to "No.5" represent the top-5 predictions at each level. The levels "Act," "Cat," and "Loc" refer to Activity, Category, and Location, respectively. The symbol $\checkmark$ indicates that the predicted label matches the ground truth at that level.}
  \vspace{-1.5ex} 
  \resizebox{0.9\textwidth}{!}{
  \fontsize{30}{36}\selectfont
  \begin{tabular}{l|c|ccccc|ccccc} 
  \toprule
  \multirow{2}{*}{Level} & \multirow{2}{*}{\begin{tabular}[c]{@{}c@{}}Ground\\ Truth\end{tabular}} & \multicolumn{5}{c|}{Ours} & \multicolumn{5}{c}{Baseline} \\ 
  \cmidrule(lr){3-7} \cmidrule(lr){8-12} 
  & & No.1 & No.2 & No.3 & No.4 & No.5 & No.1 & No.2 & No.3 & No.4 & No.5 \\
  \midrule                           
  Need  & Safety    & $\checkmark$ & $\checkmark$  & $\checkmark$  & Physiological  & Social    & Physiological & Social & Physiological & $\checkmark$ & Physiological \\
  Act   & Community & $\checkmark$ & $\checkmark$  & $\checkmark$  & Dining  & Entertainment    & Dining  & Entertainment & Retail  & $\checkmark$  & Dining   \\  
  Cat   & Home      & $\checkmark$ & School   & $\checkmark$  & Restaurant  & Multiplex         & Restaurant & Multiplex &Store &$\checkmark$ &Restaurant \\ 
  Loc   & 7121      & $\checkmark$ & 3250     & 7163          & 2765        & 77                & 2765 & 77 & 2561 & $\checkmark$ & 3503 \\
  \bottomrule                
  \end{tabular}
  }
  \vspace{-1ex} 
  \label{tab:case}
\end{table*}

\subsection{Ablation Study}  
We conduct the following ablation studies to evaluate the impact of various components of ALOHA:
(1) \textbf{w/o Exploitation}: Removes the hierarchical structure, retaining only the Gumbel disturbance.  
(2) \textbf{w/o N}: Removes the need level from the hierarchy.  
(3) \textbf{w/o N\&A}: Removes both the need and activity levels.
(4) \textbf{w/o Exploration}: Removes Gumbel disturbance and node-wise adaptive weights.  
(5) \textbf{w/o Gumbel}: Removes only the Gumbel disturbance.  
(6) \textbf{w/o Adaptive}: Removes only node-wise adaptive weights.  

As shown in Figure~\ref{fig:ablation}, using M@1 as an example, ``w/o Exploitation'' causes the largest performance drop on the more long-tailed JKT dataset, highlighting the importance of higher-level semantic exploitation for long-tailed mobility prediction; On KLP the greatest decline occurs with ``w/o Gumbel'', indicating that exploration among semantically related locations is also critical. Removing hierarchical layers generally degrades performance—especially on KLP dataset—and ``w/o N\&A'' hurts more than ``w/o N'', confirming the necessity of the four-level hierarchy. Between the two exploration mechanisms, ``w/o Gumbel'' produces a slightly larger drop than ``w/o Adaptive''. These dataset-specific trends, which vary slightly across different metrics, underscore the necessity of considering multiple evaluation metrics when selecting optimal models.

\subsection{Case Study}

We conduct a case study comparing ALOHA with the best baseline (STHGCN) on long-tailed location prediction over the JKT dataset. The true label is a tail location with only twenty-seven visits. The coarse-grained results are derived from transition matrices based on location-level predictions.
As shown in Table~\ref{tab:case}, ALOHA predicts the true label at rank one, while STHGCN fails to include it among the top three predictions. ALOHA also achieves high coarse-grained accuracy, where its top three predictions align at both need and activity levels, improving prediction diversity and reliability. In contrast, the baseline prioritizes popular locations (\eg restaurant) and overlooks rare ones. These results highlight ALOHA's effectiveness in identifying long-tailed locations and underscore the baseline's limitations in producing diverse predictions, demonstrating the benefit of hierarchical modeling.
We further define correctness by top-1 location matching and consider four scenarios: CC (both correct), WW (both wrong), CW (ours correct/baseline wrong), and WC (ours wrong/baseline correct). CW occurs 95 times versus 24 for WC. Even in WW, our model improves category prediction by 7.35\%, confirming ALOHA's strength in capturing hierarchical semantics for long-tailed mobility prediction.

\section{Conclusion}
\label{sec:conclusion}

In this paper, we introduce ALOHA, an innovative architecture-agnostic plugin designed to mitigate long-tailed bias in human mobility prediction. Grounded in Maslow's theory of human motivation, ALOHA utilizes hierarchical tree structures to ensure balanced optimization across head and tail locations.
Through an automatic pipeline leveraging LLMs and CoT prompts, ALOHA constructs city-tailored hierarchies with minimal human validation. 
The adaptive hierarchical loss incorporates Gumbel disturbance and adaptive weights to effectively exploit and explore of common knowledge patterns.
Through extensive experiments across a variety of baselines, ALOHA demonstrates superior performance, achieving up to 16.59\% improvement over advanced mobility prediction methods. Our results showcase ALOHA's robustness, efficiency and adjustment mechanism.
In future work, we aim to extend ALOHA by incorporating factors like natural disasters and large-scale events, further improving its prediction accuracy and adaptability in dynamic environments.

\begin{acks}
  This work is supported by the Zhejiang Province “JianBingLingYan+X” Research and Development Plan (2025C02020).
\end{acks}

\clearpage
\bibliographystyle{ACM-Reference-Format}
\bibliography{ref}

\appendix

\clearpage
\section{Supplement to Methodology} 

\subsection{Prompts} \label{sec:prompts}
The designed Chain-of-Thought (CoT) prompts used to generate $T (\text{Category} \rightarrow \text{Activity})$ and $T (\text{Activity} \rightarrow \text{Need})$ are as follows.

\begin{figure}[!h]
    \vspace{-3ex}                  
    \centering
    \includegraphics[width=0.5\textwidth]{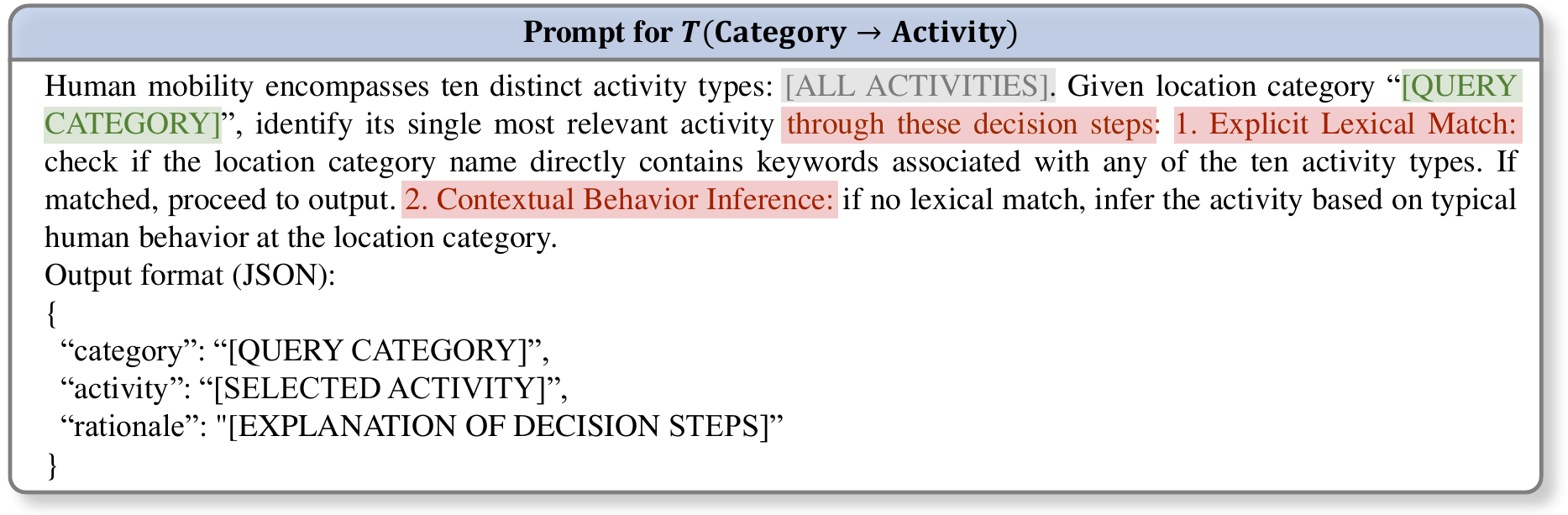} \\
    \vspace{0.6ex}
    \includegraphics[width=0.5\textwidth]{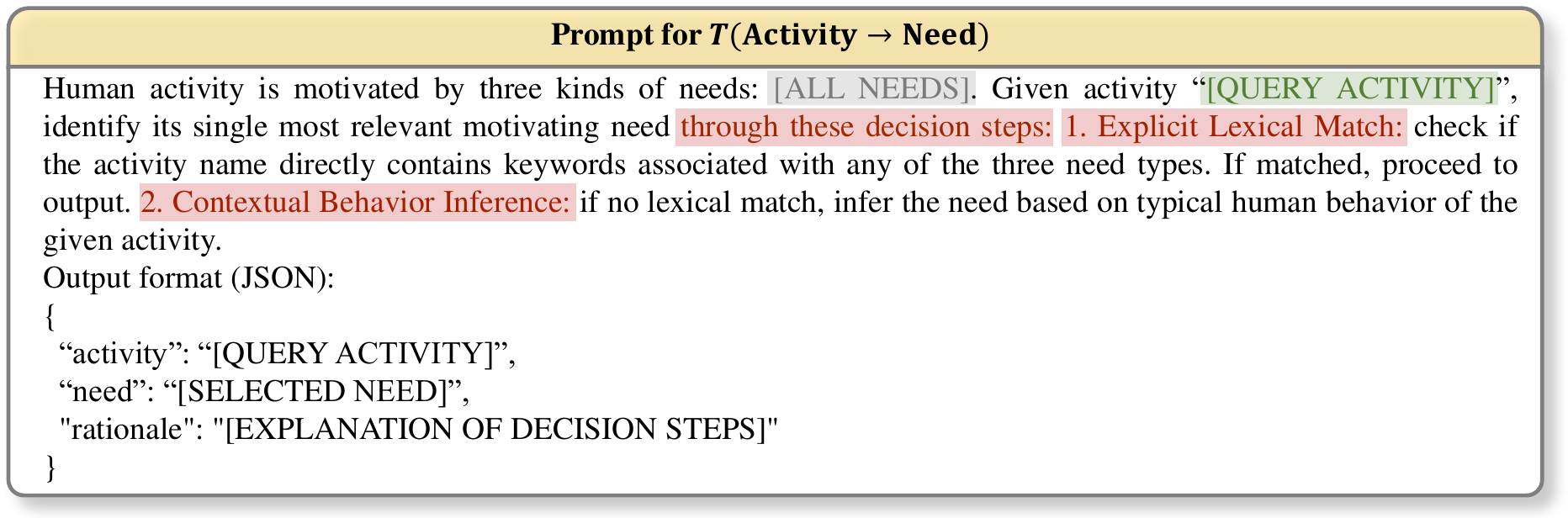}
    \vspace{-4ex}                  
    \caption{CoT prompts for relations of adjacent-level labels.} 
    \vspace{-3.5ex}                  
    \label{fig:prompts}
\end{figure}

\subsection{Evaluation for LLM-generated Mappings} \label{sec:annotation}
We establish relations between adjacent-level labels using LLMs and CoT prompts. To reduce noise, three expert annotators—two senior Ph.D. candidates and one postdoctoral researcher in psychology and behavioral sciences—verify the results. Following annotation guidelines~\cite{stab2014annotating} of Natural Language Processing (NLP), if disagreements arise, another expert is consulted for discussion. This process ensures reliable category → activity and activity → need level relations while minimizing manual effort and maintaining accuracy.

\subsection{Gradient Analysis}
\label{sec:gradient_appendix}

We analyze the gradient of our Adaptive Hierarchical Loss (AHL) with respect to logits in the location classification layer, focusing on a single prediction for clarity.
\subsubsection{Initial Probability Derivative}

We denote the logit of label \(y_j^H\) as \(z_j\) and define the initial probability for any leaf node class as:
\begin{equation}
p_0(y_i^H) = \frac{\exp\left(\frac{z_i + g_i}{\tau}\right)}{\sum_{k=1}^{C^H} \exp\left(\frac{z_k + g_k}{\tau}\right)}.
\end{equation}
Since \(\mathbf{z} = \sum_{k=1}^{C^H} \exp\left(\frac{z_k + g_k}{\tau}\right)\), we can differentiate \(p_0(y_i^H)\) with respect to \(z_j\) using the quotient rule:
\begin{equation}
\begin{aligned}
\frac{\partial p_0(y_i^H)}{\partial z_j} 
&= p_0(y_i^H)(\delta_{ij} - p_0(y_i^H)),
\end{aligned}
\end{equation}
where $\delta_{ij}$ is the Kronecker delta, equal to 1 if $i=j$ and 0 otherwise.
\subsubsection{Hierarchical Conditional Probability Derivative}

Let $y_j^h$ denote a label at the $h$-th level. For simplicity, we define $S_i^h$ as the sum of probabilities of all leaf nodes under the non-leaf node $y_i^h$. The conditional probability of any class $y_i^h$ at level $h$ is then:
\begin{equation}
p(y_i^h \mid y_i^{h-1}) = \frac{S_i^h}{S_i^{h-1}}, \quad S_i^h = \sum_{y_k^H \in \text{Leaves}(y_i^h)} p_0(y_k^H).
\end{equation}
The derivative of this conditional probability with respect to \(z_j\) is:
\begin{equation}
\frac{\partial p(y_i^h \mid y_i^{h-1})}{\partial z_j} = \frac{\partial S_i^h}{\partial z_j} \cdot \frac{1}{S_i^{h-1}} - \frac{S_i^h}{(S_i^{h-1})^2} \cdot \frac{\partial S_i^{h-1}}{\partial z_j}.
\end{equation}
Here, we expand the derivative of \(S_i^h\) with respect to \(z_j\):
\begin{equation}
    \frac{\partial S_i^h}{\partial z_j} = \sum_{y_k^H \in \operatorname{Leaves}(y_i^h)} \frac{\partial p_0(y_k^H)}{\partial z_j}.
\end{equation}
We need to consider two situations. If \(y_i^H \in \text{Leaves}(y_i^h)\), then:
\begin{equation}
    \begin{aligned}
        \frac{\partial S_i^h}{\partial z_j} 
        & = p_0(y_i^H)(1 - S_i^h).
    \end{aligned}
\end{equation}
Otherwise:
\begin{equation}
    \frac{\partial S_i^h}{\partial z_j} = -S_i^h p_0(y_i^H).
\end{equation}

\subsubsection{Adaptive Hierarchical Loss Function Derivative}

The conditional probability \(p(y_i^h \mid y_i^{h-1})\) depends on \(z_i\) if and only if the node \(y_i^h\) is an ancestor of the leaf node \(y_i^H\). Let the ancestral path of \(y_i^H\) be \(\{y_i^1, y_i^2, \ldots, y_i^H\}\). The loss function is defined as:
\begin{equation}
\ell_{\text{hier}} = -\sum_{h=1}^H {w}^h_i \log p(y_i^h \mid y_i^{h-1}).
\end{equation}
The derivative of the loss function with respect to \(z_j\) is given by:
\begin{equation}
\begin{aligned}
\frac{\partial \ell_{\text{hier}}}{\partial z_j} 
&= -\sum_{h=1}^H {w}^h_i \left[\frac{1}{S_i^h} \cdot \frac{\partial S_i^h}{\partial z_j} - \frac{1}{S_{h-1}} \cdot \frac{\partial S_{h-1}}{\partial z_j}\right] \\
&= p_0(y_i^H) \left[\sum_{h=1}^{H-1} \frac{{w}_i^{h+1} - {w}^h_i}{S_i^h} + {w}^1_i\right] - {w}_i^H \\
&= p_0(y_i^H) \cdot A - {w}_i^H,
\end{aligned}
\end{equation}
where $A = \sum_{h=1}^{H-1} \frac{{w}_i^{h+1} - {w}^h_i}{S_i^h} + {w}^1_i$. This gradient analysis reveals how changes in logits affect probabilities and the loss across the hierarchy, providing key insights for understanding and optimizing the model's training dynamics.

\begin{figure}[!ht]
    \centering	
    \includegraphics[width=0.4\textwidth]{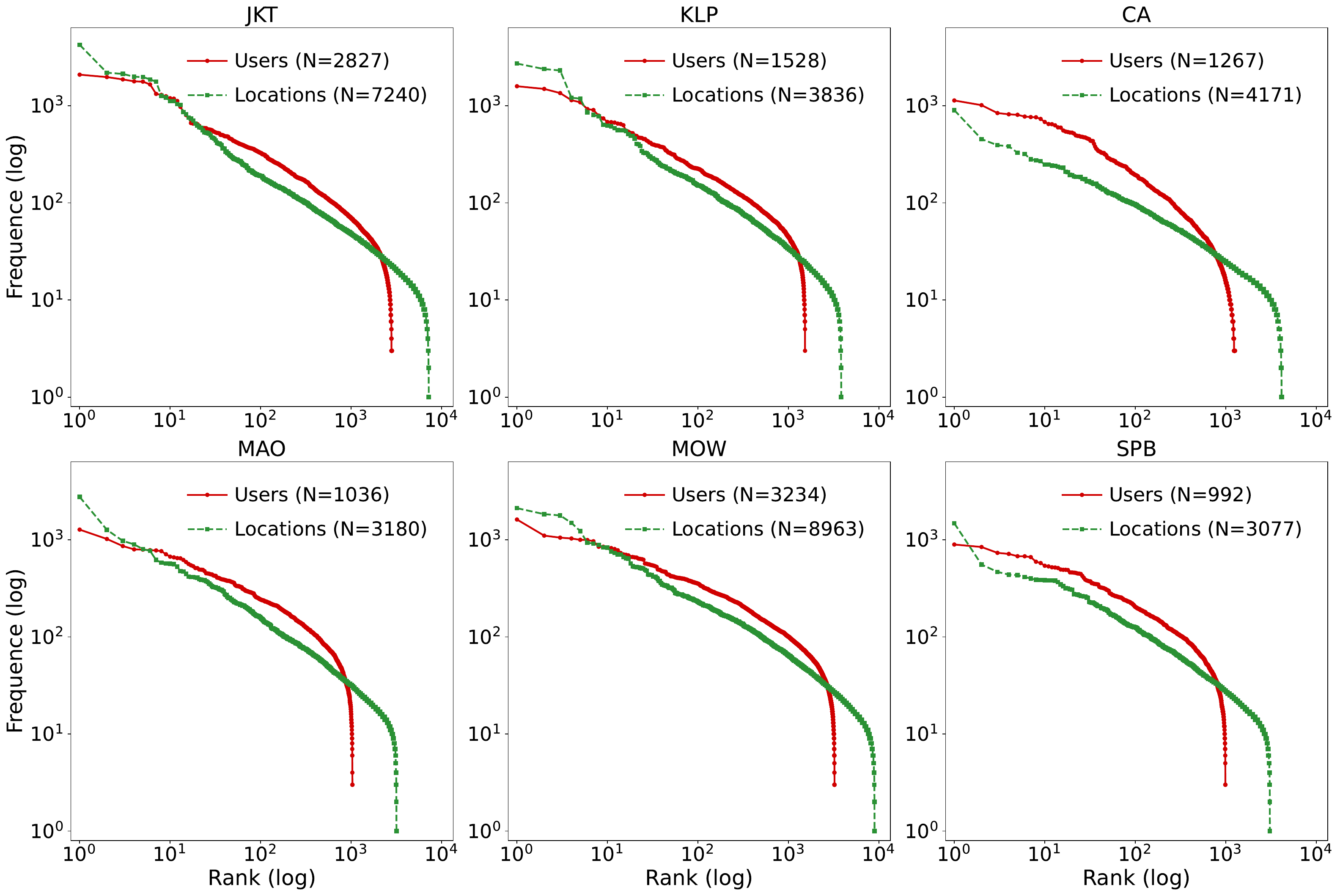}
    \vspace{-3ex}                  
    \caption{Distribution of users and locations on six datasets.}
    \label{fig:distribution}
    \vspace{-2ex}                 
\end{figure}

\begin{table*}[!t]
  \caption{M@$k$ and N@$k$ on MAO, MOW, and SPB datasets. 
  GFlash denotes Graph-Flashback. 
  Best and second-best scores are marked in blue and light blue. Improv. is the relative percentage improvement.}
  \label{tab:main_appendix}
  \vspace{-1ex}

\resizebox{0.91\textwidth}{!}{
  \begin{tabular}{l|ccccc|ccccc|ccccc}
    \toprule
    Dataset &
  \multicolumn{5}{c|}{MAO} &
  \multicolumn{5}{c|}{MOW} &
  \multicolumn{5}{c}{SPB} \\ \hline
Metric &
  M@1 &
  M@5 &
  M@10 &
  N@5 &
  N@10 &
  M@1 &
  M@5 &
  M@10 &
  N@5 &
  N@10 &
  M@1 &
  M@5 &
  M@10 &
  N@5 &
  N@10 \\ \hline
GFlash &
  \cellcolor[HTML]{E4E8F5}0.2093 &
  0.3175 &
  0.3315 &
  0.3631 &
  0.3967 &
  \cellcolor[HTML]{E4E8F5}0.1752 &
  \cellcolor[HTML]{E4E8F5}0.2644 &
  \cellcolor[HTML]{E4E8F5}0.2737 &
  \cellcolor[HTML]{E4E8F5}0.3017 &
  \cellcolor[HTML]{E4E8F5}0.3240 &
  0.2019 &
  0.2959 &
  0.3061 &
  0.3338 &
  0.3583 \\
+Focal &
  0.1962 &
  0.3107 &
  0.3249 &
  0.3588 &
  0.3923 &
  0.1679 &
  0.2618 &
  0.2711 &
  0.3001 &
  0.3225 &
  \cellcolor[HTML]{E4E8F5}0.2128 &
  \cellcolor[HTML]{E4E8F5}0.3011 &
  \cellcolor[HTML]{E4E8F5}0.3105 &
  0.3367 &
  0.3592 \\
+CB &
  0.1858 &
  0.2824 &
  0.2951 &
  0.3243 &
  0.3548 &
  0.1576 &
  0.2414 &
  0.2504 &
  0.2764 &
  0.2983 &
  0.2001 &
  0.2889 &
  0.2964 &
  0.3250 &
  0.3430 \\
+IB &
  0.1816 &
  0.2793 &
  0.2917 &
  0.3218 &
  0.3514 &
  0.1428 &
  0.2290 &
  0.2394 &
  0.2634 &
  0.2882 &
  0.1658 &
  0.2599 &
  0.2709 &
  0.2975 &
  0.3238 \\
+LA &
  0.1882 &
  0.2879 &
  0.2998 &
  0.3311 &
  0.3598 &
  0.1654 &
  0.2510 &
  0.2608 &
  0.2862 &
  0.3096 &
  0.1977 &
  0.2898 &
  0.2989 &
  0.3274 &
  0.3489 \\
+BLV &
  0.2093 &
  \cellcolor[HTML]{E4E8F5}0.3196 &
  \cellcolor[HTML]{E4E8F5}0.3338 &
  \cellcolor[HTML]{E4E8F5}0.3657 &
  \cellcolor[HTML]{E4E8F5}0.3994 &
  0.1719 &
  0.2630 &
  0.2729 &
  0.2998 &
  0.3236 &
  0.2098 &
  0.3003 &
  0.3099 &
  \cellcolor[HTML]{E4E8F5}0.3383 &
  \cellcolor[HTML]{E4E8F5}0.3611 \\
+Ours &
  \cellcolor[HTML]{B4C6E7}0.2182 &
  \cellcolor[HTML]{B4C6E7}0.3344 &
  \cellcolor[HTML]{B4C6E7}0.3456 &
  \cellcolor[HTML]{B4C6E7}0.3838 &
  \cellcolor[HTML]{B4C6E7}0.4106 &
  \cellcolor[HTML]{B4C6E7}0.1922 &
  \cellcolor[HTML]{B4C6E7}0.2800 &
  \cellcolor[HTML]{B4C6E7}0.2887 &
  \cellcolor[HTML]{B4C6E7}0.3159 &
  \cellcolor[HTML]{B4C6E7}0.3365 &
  \cellcolor[HTML]{B4C6E7}0.2363 &
  \cellcolor[HTML]{B4C6E7}0.3210 &
  \cellcolor[HTML]{B4C6E7}0.3313 &
  \cellcolor[HTML]{B4C6E7}0.3562 &
  \cellcolor[HTML]{B4C6E7}0.3810 \\
Improv. &
  4.25\% &
  4.63\% &
  3.54\% &
  4.95\% &
  2.80\% &
  9.70\% &
  5.90\% &
  5.48\% &
  4.71\% &
  3.86\% &
  11.04\% &
  6.61\% &
  6.70\% &
  5.29\% &
  5.51\% \\ \hline
STHGCN &
  \cellcolor[HTML]{E4E8F5}0.2281 &
  \cellcolor[HTML]{E4E8F5}0.3406 &
  \cellcolor[HTML]{E4E8F5}0.3522 &
  \cellcolor[HTML]{E4E8F5}0.3869 &
  \cellcolor[HTML]{E4E8F5}0.4143 &
  \cellcolor[HTML]{E4E8F5}0.1815 &
  0.2629 &
  0.2722 &
  0.2963 &
  0.3186 &
  0.1844 &
  0.2754 &
  0.2846 &
  0.3119 &
  0.3338 \\
+Focal &
  0.2248 &
  0.3356 &
  0.3470 &
  0.3831 &
  0.4103 &
  0.1772 &
  0.2623 &
  0.2707 &
  0.2968 &
  0.3171 &
  0.1935 &
  0.2823 &
  0.2915 &
  0.3184 &
  0.3408 \\
+CB &
  0.2013 &
  0.3031 &
  0.3125 &
  0.3470 &
  0.3693 &
  0.1643 &
  0.2405 &
  0.2484 &
  0.2708 &
  0.2897 &
  \cellcolor[HTML]{E4E8F5}0.1977 &
  \cellcolor[HTML]{E4E8F5}0.2832 &
  0.2911 &
  0.3182 &
  0.3376 \\
+IB &
  0.1962 &
  0.3012 &
  0.3110 &
  0.3446 &
  0.3680 &
  0.1553 &
  0.2268 &
  0.2339 &
  0.2551 &
  0.2723 &
  0.1875 &
  0.2769 &
  0.2861 &
  0.3131 &
  0.3351 \\
+LA &
  0.2173 &
  0.3225 &
  0.3321 &
  0.3676 &
  0.3904 &
  0.1665 &
  0.2426 &
  0.2512 &
  0.2741 &
  0.2947 &
  0.1887 &
  0.2832 &
  \cellcolor[HTML]{E4E8F5}0.2929 &
  \cellcolor[HTML]{E4E8F5}0.3195 &
  \cellcolor[HTML]{E4E8F5}0.3424 \\
+BLV &
  0.2262 &
  0.3392 &
  0.3504 &
  0.3862 &
  0.4128 &
  0.1784 &
  \cellcolor[HTML]{E4E8F5}0.2639 &
  \cellcolor[HTML]{E4E8F5}0.2724 &
  \cellcolor[HTML]{E4E8F5}0.2989 &
  \cellcolor[HTML]{E4E8F5}0.3196 &
  0.1869 &
  0.2729 &
  0.2823 &
  0.3076 &
  0.3307 \\
+Ours &
  \cellcolor[HTML]{B4C6E7}0.2426 &
  \cellcolor[HTML]{B4C6E7}0.3510 &
  \cellcolor[HTML]{B4C6E7}0.3611 &
  \cellcolor[HTML]{B4C6E7}0.3962 &
  \cellcolor[HTML]{B4C6E7}0.4205 &
  \cellcolor[HTML]{B4C6E7}0.1846 &
  \cellcolor[HTML]{B4C6E7}0.2705 &
  \cellcolor[HTML]{B4C6E7}0.2786 &
  \cellcolor[HTML]{B4C6E7}0.3051 &
  \cellcolor[HTML]{B4C6E7}0.3248 &
  \cellcolor[HTML]{B4C6E7}0.2007 &
  \cellcolor[HTML]{B4C6E7}0.2857 &
  \cellcolor[HTML]{B4C6E7}0.2953 &
  \cellcolor[HTML]{B4C6E7}0.3204 &
  \cellcolor[HTML]{B4C6E7}0.3436 \\
Improv. &
  6.36\% &
  3.05\% &
  2.53\% &
  2.40\% &
  1.50\% &
  1.71\% &
  2.50\% &
  2.28\% &
  2.07\% &
  1.63\% &
  1.52\% &
  0.88\% &
  0.82\% &
  0.28\% &
  0.35\% \\ \hline
MCLP &
  \cellcolor[HTML]{E4E8F5}0.2163 &
  \cellcolor[HTML]{E4E8F5}0.3285 &
  \cellcolor[HTML]{E4E8F5}0.3406 &
  \cellcolor[HTML]{E4E8F5}0.3744 &
  \cellcolor[HTML]{E4E8F5}0.4036 &
  0.1770 &
  0.2616 &
  0.2714 &
  0.2964 &
  0.3197 &
  0.2031 &
  0.2925 &
  0.3011 &
  0.3298 &
  0.3503 \\
+Focal &
  0.2088 &
  0.3250 &
  0.3358 &
  0.3726 &
  0.3983 &
  0.1761 &
  \cellcolor[HTML]{E4E8F5}0.2648 &
  \cellcolor[HTML]{E4E8F5}0.2735 &
  \cellcolor[HTML]{E4E8F5}0.3009 &
  \cellcolor[HTML]{E4E8F5}0.3216 &
  0.2025 &
  0.2938 &
  0.3025 &
  0.3312 &
  0.3521 \\
+CB &
  0.1957 &
  0.2981 &
  0.3097 &
  0.3419 &
  0.3696 &
  0.1641 &
  0.2482 &
  0.2567 &
  0.2830 &
  0.3033 &
  0.1971 &
  0.2863 &
  0.2949 &
  0.3217 &
  0.3422 \\
+IB &
  0.2004 &
  0.2863 &
  0.2942 &
  0.3214 &
  0.3395 &
  0.1757 &
  0.2426 &
  0.2480 &
  0.2690 &
  0.2817 &
  0.1977 &
  0.2777 &
  0.2851 &
  0.3084 &
  0.3258 \\
+LA &
  0.2023 &
  0.3079 &
  0.3198 &
  0.3520 &
  0.3804 &
  \cellcolor[HTML]{E4E8F5}0.1788 &
  0.2634 &
  0.2713 &
  0.2978 &
  0.3166 &
  0.2001 &
  0.2936 &
  0.3030 &
  0.3291 &
  0.3516 \\
+BLV &
  0.2126 &
  0.3228 &
  0.3357 &
  0.3675 &
  0.3984 &
  0.1770 &
  0.2618 &
  0.2712 &
  0.2966 &
  0.3192 &
  \cellcolor[HTML]{E4E8F5}0.2086 &
  \cellcolor[HTML]{E4E8F5}0.2969 &
  \cellcolor[HTML]{E4E8F5}0.3057 &
  \cellcolor[HTML]{E4E8F5}0.3339 &
  \cellcolor[HTML]{E4E8F5}0.3554 \\
+Ours &
  \cellcolor[HTML]{B4C6E7}0.2290 &
  \cellcolor[HTML]{B4C6E7}0.3403 &
  \cellcolor[HTML]{B4C6E7}0.3511 &
  \cellcolor[HTML]{B4C6E7}0.3869 &
  \cellcolor[HTML]{B4C6E7}0.4128 &
  \cellcolor[HTML]{B4C6E7}0.1889 &
  \cellcolor[HTML]{B4C6E7}0.2725 &
  \cellcolor[HTML]{B4C6E7}0.2819 &
  \cellcolor[HTML]{B4C6E7}0.3076 &
  \cellcolor[HTML]{B4C6E7}0.3300 &
  \cellcolor[HTML]{B4C6E7}0.2140 &
  \cellcolor[HTML]{B4C6E7}0.3049 &
  \cellcolor[HTML]{B4C6E7}0.3135 &
  \cellcolor[HTML]{B4C6E7}0.3416 &
  \cellcolor[HTML]{B4C6E7}0.3620 \\
Improv. &
  5.87\% &
  3.59\% &
  3.08\% &
  3.34\% &
  2.28\% &
  5.65\% &
  2.91\% &
  3.07\% &
  2.23\% &
  2.61\% &
  2.59\% &
  2.69\% &
  2.55\% &
  2.31\% &
  1.86\% \\ \hline
Diff-POI &
  \cellcolor[HTML]{E4E8F5}0.1455 &
  \cellcolor[HTML]{E4E8F5}0.2148 &
  \cellcolor[HTML]{E4E8F5}0.2230 &
  \cellcolor[HTML]{E4E8F5}0.2439 &
  \cellcolor[HTML]{E4E8F5}0.2638 &
  \cellcolor[HTML]{E4E8F5}0.1208 &
  \cellcolor[HTML]{E4E8F5}0.1781 &
  \cellcolor[HTML]{E4E8F5}0.1841 &
  \cellcolor[HTML]{E4E8F5}0.2020 &
  \cellcolor[HTML]{E4E8F5}0.2165 &
  0.1332 &
  0.2070 &
  0.2133 &
  0.2357 &
  0.2509 \\
+Focal &
  0.1394 &
  0.2105 &
  0.2187 &
  0.2408 &
  0.2609 &
  0.1159 &
  0.1723 &
  0.1781 &
  0.1957 &
  0.2096 &
  0.1368 &
  \cellcolor[HTML]{E4E8F5}0.2089 &
  \cellcolor[HTML]{E4E8F5}0.2167 &
  \cellcolor[HTML]{E4E8F5}0.2390 &
  \cellcolor[HTML]{E4E8F5}0.2576 \\
+CB &
  0.1300 &
  0.1967 &
  0.2037 &
  0.2252 &
  0.2416 &
  0.1128 &
  0.1668 &
  0.1715 &
  0.1880 &
  0.1992 &
  0.1272 &
  0.1958 &
  0.2024 &
  0.2235 &
  0.2390 \\
+IB &
  0.1450 &
  0.1790 &
  0.1827 &
  0.1927 &
  0.2015 &
  0.0871 &
  0.1029 &
  0.1058 &
  0.1100 &
  0.1170 &
  0.1200 &
  0.1502 &
  0.1539 &
  0.1624 &
  0.1714 \\
+LA &
  0.1314 &
  0.1982 &
  0.2073 &
  0.2273 &
  0.2490 &
  0.1045 &
  0.1632 &
  0.1681 &
  0.1870 &
  0.1987 &
  0.1290 &
  0.1968 &
  0.2043 &
  0.2244 &
  0.2427 \\
+BLV &
  0.1427 &
  0.2138 &
  0.2226 &
  0.2429 &
  0.2640 &
  0.1166 &
  0.1699 &
  0.1767 &
  0.1916 &
  0.2078 &
  \cellcolor[HTML]{E4E8F5}0.1374 &
  0.2026 &
  0.2113 &
  0.2284 &
  0.2488 \\
+Ours &
  \cellcolor[HTML]{B4C6E7}0.1694 &
  \cellcolor[HTML]{B4C6E7}0.2343 &
  \cellcolor[HTML]{B4C6E7}0.2420 &
  \cellcolor[HTML]{B4C6E7}0.2613 &
  \cellcolor[HTML]{B4C6E7}0.2800 &
  \cellcolor[HTML]{B4C6E7}0.1321 &
  \cellcolor[HTML]{B4C6E7}0.1870 &
  \cellcolor[HTML]{B4C6E7}0.1927 &
  \cellcolor[HTML]{B4C6E7}0.2098 &
  \cellcolor[HTML]{B4C6E7}0.2236 &
  \cellcolor[HTML]{B4C6E7}0.1555 &
  \cellcolor[HTML]{B4C6E7}0.2238 &
  \cellcolor[HTML]{B4C6E7}0.2310 &
  \cellcolor[HTML]{B4C6E7}0.2509 &
  \cellcolor[HTML]{B4C6E7}0.2678 \\
Improv. &
  16.43\% &
  9.08\% &
  8.52\% &
  7.13\% &
  6.06\% &
  9.35\% &
  5.00\% &
  4.67\% &
  3.86\% &
  3.28\% &
  13.17\% &
  7.13\% &
  6.60\% &
  4.98\% &
  3.96\% \\ \hline
MELT &
  0.1940 &
  0.3111 &
  0.3221 &
  0.3616 &
  0.3877 &
  0.1348 &
  0.2189 &
  0.2283 &
  0.2539 &
  0.2765 &
  0.1630 &
  0.2562 &
  0.2645 &
  0.2933 &
  0.3136 \\
LoTNext &
  0.2098 &
  0.3118 &
  0.3238 &
  0.3555 &
  0.384 &
  0.1837 &
  0.2626 &
  0.2713 &
  0.2955 &
  0.3159 &
  0.1911 &
  0.2652 &
  0.2736 &
  0.2955 &
  0.3159 \\ \bottomrule
\end{tabular}
}
\vspace{-1ex}
\end{table*}

\section{Supplement to Experiments} \label{sec:exp_appendix}
\subsection{Datasets}
\label{sec:statistics}
We analyze the frequency distribution of users and locations across six human mobility datasets (Figure~\ref{fig:distribution}). Both user and location frequencies follow a long-tailed distribution where a few users and locations dominate interactions. In cities like KLP and MOW, the user and location curves are similar, indicating more uniform exploration. In contrast, cities like MAO and SPB show a sharp drop in interaction frequency, reflecting a concentration of visits to a few popular locations, with many remaining under-visited.

\subsection{Baselines}
\label{sec:baseline}
We evaluate ALOHA on four advanced mobility prediction models: 
Graph-Flashback~\cite{rao2022graph}, which leverages RNNs with user-POI knowledge and transition graphs; 
STHGCN~\cite{yan2023spatio}, which employs hypergraph Transformers with spatiotemporal encoding; 
MCLP~\cite{sun2024going}, which integrates topic modeling, time embeddings, and Transformers; 
and Diff-POI~\cite{qin2023diffusion}, which combines dual graph encoders with diffusion-based sampling. 
We further compare against two categories of long-tailed learning methods: 
\textbf{Plugins}, including:  
Focal Loss (Focal)~\cite{lin2017focal}, which reshapes cross-entropy to down-weight easy examples and focus training on hard instances;  
Class-Balanced Loss (CB)~\cite{cui2019class}, which re-weights classes using the effective number of samples to alleviate imbalance;  
IB Loss (IB)~\cite{park2021influence}, which balances training by reducing the influence of samples that bias the decision boundary;  
Logit Adjustment (LA)~\cite{menonlong}, which adjusts logits post-hoc based on label frequencies to mitigate skewed predictions;  
and BLV~\cite{wang2023balancing}, which introduces category-wise variation during training to enhance balance in semantic segmentation.  
\textbf{Frameworks}, such as MELT~\cite{kim2023melt}, which adopts bilateral branches with curriculum learning, 
and LoTNext~\cite{cui2019class}, which combines graph adjustment, loss modification, and auxiliary tasks for sequential recommendation.

\begin{figure*}[!ht]
  \centering	
  \includegraphics[width=\textwidth]{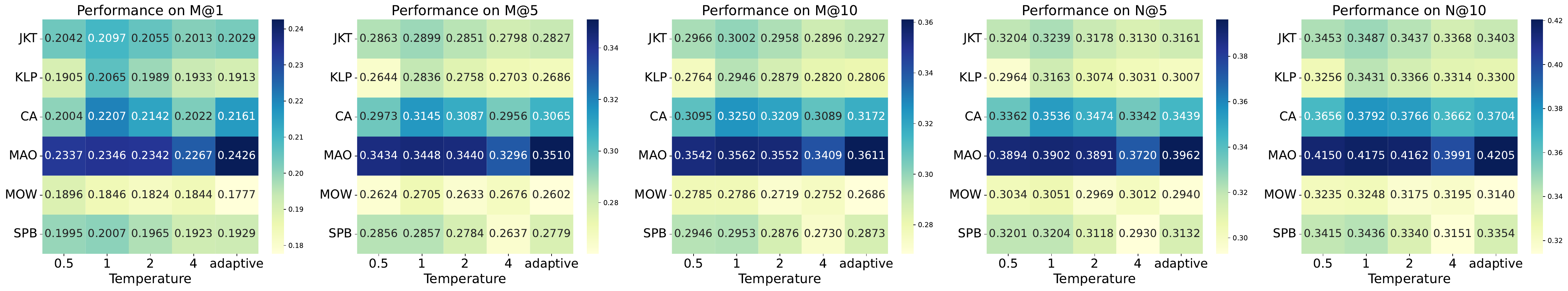}
  \vspace{-2.5ex}
  \caption{The impact of parameter $\tau$ in our proposed ALOHA using STHGCN across six datasets.}
  \label{fig:hyper_graph}
\end{figure*}

\subsection{Implementation Details}
\label{sec:implementation}
All experiments are conducted on a single RTX A6000 GPU with Adam optimizer and seed 42. Graph-Flashback, STHGCN, and MCLP use 128-d embeddings, while Diff-POI uses 64-d with dual graph encoders ($\alpha=0.5,\gamma=0.2,\lambda=10^{-3},\delta_s=\delta_t=256$). STHGCN employs a 2-layer hypergraph convolution ($\beta=0.5$, 4 heads), and MCLP integrates 400 LDA topics with a 2-layer Transformer (4 heads). For Gumbel-Softmax in ALOHA, temperatures are searched over $\{0.5,1,2,4\}$ and an adaptive annealing strategy from 2 to 0.5. Hierarchical weights are initialized as $\{1,0.75,0.5,0.25\}$. Plugin baselines include Focal ($\gamma=2$), CB ($\beta=0.9999$), IB ($\epsilon=0.001$), LA ($\tau=1$), and BLV ($\sigma=4$). 
Framework baselines include LoTNext searched over hidden sizes $\{32,128\}$, batch sizes $\{16,32,64\}$, and learning rates $\{1\text{e-4},1\text{e-3}\}$, and MELT with hidden size 128, max length 200, $\lambda_U=\lambda_I=0.1$, and 2 Transformer layers.

\subsection{Prediction Performance}
\label{sec:main_table_appendix}

Results on the remaining three datasets are summarized in Table~\ref{tab:main_appendix}. Our method consistently outperforms baselines across all five metrics, demonstrating robustness and generalizability across backbone architectures. On MAO, ALOHA improves M@1 by 16.43\% over the second-best Diff-POI. On MAO, MOW, and SPB, most plugins underperform compared to the base models, highlighting their limitations for complex spatiotemporal mobility prediction. On the sparse SPB dataset, BLV achieves the second-best performance on MCLP, reflecting better adaptability than other vision-derived plugins. Notably, the Diff-POI backbone performs poorly across datasets, indicating challenges in long-tailed prediction under sparse data. Finally, MELT performs worse than basic backbones due to differences in scale and strategy between recommendation system and human mobility prediction tasks.

\begin{table}[!h]
  \caption{Prediction performance under noisy hierarchical mappings on the KLP dataset, where ``N'', ``A'', and ``NA'' denote noise in activity $\rightarrow$ need, category $\rightarrow$ activity, and both.}
  \vspace{-1.8ex}
  \resizebox{0.36\textwidth}{!}{
  \begin{tabular}{l|ccccc}
      \toprule
      \textbf{Metric} & \textbf{M@1} & \textbf{M@5} & \textbf{M@10} & \textbf{N@5} & \textbf{N@10} \\
      \midrule
      N               & 0.1969       & 0.2787       & 0.2894        & 0.3135       & 0.3396        \\
      A               & 0.1901       & 0.2704       & 0.2829        & 0.3040       & 0.3340        \\
      NA              & 0.1917       & 0.2682       & 0.2792        & 0.3003       & 0.3264        \\
      Ours            & 0.2065       & 0.2836       & 0.2946        & 0.3163       & 0.3431        \\
      \bottomrule
  \end{tabular}
  }
  \vspace{-3ex}
  \label{tab:noisy_klp}
\end{table}

\begin{table}[!h]
  \caption{Prediction performance under noisy hierarchical mappings on the CA dataset.}
  \vspace{-1.8ex}
  \resizebox{0.36\textwidth}{!}{
  \begin{tabular}{l|ccccc}
      \toprule
      \textbf{Metric} & \textbf{M@1} & \textbf{M@5} & \textbf{M@10} & \textbf{N@5} & \textbf{N@10} \\
      \midrule
      N               & 0.2004       & 0.2993       & 0.3121        & 0.3394       & 0.3705        \\
      A               & 0.2105       & 0.3045       & 0.3173        & 0.3426       & 0.3733        \\
      NA              & 0.2151       & 0.3095       & 0.3223        & 0.3491       & 0.3796        \\
      Ours            & 0.2207       & 0.3145       & 0.3250        & 0.3536       & 0.3792        \\ 
      \bottomrule
  \end{tabular}
  }
  \vspace{-2ex}
  \label{tab:noisy_ca}
\end{table}

\begin{table}[!h]
  \caption{Prediction performance under noisy hierarchical mappings on the MAO dataset.}
  \vspace{-1.6ex}
  \resizebox{0.36\textwidth}{!}{
  \begin{tabular}{l|ccccc}
      \toprule
      \textbf{Metric} & \textbf{M@1} & \textbf{M@5} & \textbf{M@10} & \textbf{N@5} & \textbf{N@10} \\
      \midrule
      N               & 0.2412       & 0.3500       & 0.3598        & 0.3961       & 0.4198        \\
      A               & 0.2323       & 0.3425       & 0.3537        & 0.3876       & 0.4144        \\
      NA              & 0.2304       & 0.3380       & 0.3484        & 0.3835       & 0.4082        \\
      Ours            & 0.2426       & 0.3510       & 0.3611        & 0.3962       & 0.4205        \\ 
      \bottomrule
  \end{tabular}
  }
  \label{tab:noisy_mao}
\end{table}

\begin{table}[!h]
  \caption{Prediction performance under noisy hierarchical mappings on the MOW dataset.}
  \vspace{-1.6ex}
  \resizebox{0.36\textwidth}{!}{
  \begin{tabular}{l|ccccc}
      \toprule
      \textbf{Metric} & \textbf{M@1} & \textbf{M@5} & \textbf{M@10} & \textbf{N@5} & \textbf{N@10} \\
      \midrule
      N               & 0.1896       & 0.2736       & 0.2818        & 0.3078       & 0.3274        \\
      A               & 0.1882       & 0.2709       & 0.2794        & 0.3043       & 0.3248        \\
      NA              & 0.1860       & 0.2693       & 0.2782        & 0.3027       & 0.3237        \\
      Ours            & 0.1846       & 0.2705       & 0.2786        & 0.3051       & 0.3248        \\
      \bottomrule
  \end{tabular}
  }
  \label{tab:noisy_mow}
\end{table}

\begin{table}[!h]
  \caption{Prediction performance under noisy hierarchical mappings on the SPB dataset.}
  \vspace{-1.6ex}
  \resizebox{0.36\textwidth}{!}{
  \begin{tabular}{l|ccccc}
      \toprule
      \textbf{Metric} & \textbf{M@1} & \textbf{M@5} & \textbf{M@10} & \textbf{N@5} & \textbf{N@10} \\
      \midrule
      N               & 0.2001       & 0.2825       & 0.2934        & 0.3165       & 0.3427        \\
      A               & 0.2025       & 0.2869       & 0.2961        & 0.3221       & 0.3444        \\
      NA              & 0.2007       & 0.2875       & 0.2966        & 0.3226       & 0.3446        \\
      Ours            & 0.2007       & 0.2857       & 0.2953        & 0.3204       & 0.3436        \\ 
      \bottomrule
  \end{tabular}
  }
  \label{tab:noisy_spb}
\end{table}

\subsection{Robustness to Noisy Hierarchy}\label{sec:noisy_appendix}
We evaluate the robustness of ALOHA by randomly corrupting 10\% of hierarchical mappings for activity $\rightarrow$ need (N), category $\rightarrow$ activity (A), and both (NA). The prediction performance under noisy hierarchical mappings on the left five datasets are presented in Tables~\ref{tab:noisy_klp}--\ref{tab:noisy_spb}, where performance remains nearly identical to the clean setting, with only minor metric variations. This demonstrates that ALOHA is robust to moderate noise in the hierarchical mappings, with its adaptive mechanisms effectively mitigating the noise's impact. 

\subsection{Hyperparameter Analysis}
Temperature $\tau$ plays a crucial role in the Gumbel disturbance of ALOHA, as it controls the smoothness of perturbed logits and rebalances predictions between head and tail locations. To assess its impact, we test four fixed values $\{0.5,1,2,4\}$ and one adaptive schedule from $2$ to $0.5$ on the strongest backbone (STHGCN) across six datasets. As shown in Figure~\ref{fig:hyper_graph}, ALOHA exhibits robust performance under different $\tau$ settings: most datasets (JKT, KLP, CA, MOW, SPB) perform best at $\tau=1$, while the adaptive strategy achieves the best results on the MOW dataset.

\end{document}